\documentclass{article}

\usepackage{microtype}
\usepackage{graphicx}
\usepackage{subcaption}
\usepackage{booktabs} 

\usepackage{hyperref}

\usepackage[accepted]{icml2026}

\usepackage{amsmath}
\usepackage{amssymb}
\usepackage{mathtools}
\usepackage{amsthm}

\usepackage[capitalize,noabbrev]{cleveref}

\theoremstyle{plain}

\theoremstyle{definition}

\theoremstyle{remark}

\usepackage[textsize=tiny]{todonotes}
\usepackage{hyperref}
\usepackage{url}
\usepackage{amsmath}
\usepackage{amssymb}
\usepackage{mathtools}
\usepackage{amsthm}
\usepackage{bm} 
\usepackage{thmtools, thm-restate}
\usepackage{tcolorbox}
\tcbuselibrary{skins}
\usepackage{hyperref}
\usepackage{pgfplots}
\usepackage{xcolor}
\usepackage{dsfont}
\usepackage{xcolor}
\usepackage{algorithm}
\usepackage{algorithmic}
\usepackage{subcaption} 
\definecolor{softbluegray}{RGB}{112,138,153}
\usepackage{enumitem}
\usepackage{placeins}

\icmltitlerunning{K-Myriad: Jump-starting reinforcement learning}

\begin{document}

\twocolumn[
  \icmltitle{K-Myriad: Jump-starting reinforcement learning\\  with unsupervised parallel agents}



  \icmlsetsymbol{equal}{*}

  \begin{icmlauthorlist}
    \icmlauthor{Vincenzo De Paola}{sch}
    \icmlauthor{Mirco Mutti}{sch_h}
    \icmlauthor{Riccardo Zamboni}{sch}
    \icmlauthor{Marcello Restelli}{sch}

  \end{icmlauthorlist}

  \icmlaffiliation{sch}{AIRLAB, Politecnico di Milano, Milan, Italy}
   \icmlaffiliation{sch_h}{Technion, Israel Intitute of Technology, Haifa, Israel}

\icmlcorrespondingauthor{Vincenzo De Paola}{vincenzo.depaola@polimi.it}

  \icmlkeywords{Machine Learning, ICML}

  \vskip 0.3in
]



\printAffiliationsAndNotice{}  

\begin{abstract}
Parallelization in Reinforcement Learning is typically employed to speed up the training of a single policy, where multiple workers collect experience from an identical sampling distribution.
This common design limits the potential of parallelization by neglecting the advantages of diverse exploration strategies.
We propose \textbf{K-Myriad}, a scalable and unsupervised method that maximizes the collective state entropy induced by a population of parallel policies.
By cultivating a portfolio of specialized exploration strategies, K-Myriad provides a robust initialization for Reinforcement Learning, leading to both higher training efficiency and the discovery of heterogeneous solutions. Experiments on high-dimensional continuous control tasks, with large-scale parallelization, demonstrate that K-Myriad can learn a broad set of distinct policies, highlighting its effectiveness for collective exploration and paving the way towards novel parallelization strategies.
\end{abstract}

\section{Introduction}
\label{sec:introduction}

Despite its recent successes, most notably for finetuning of large-scale language models and reasoning models~\citep{christiano2017deep,guo2025deepseek}, Reinforcement Learning~\citep[RL,][]{sutton} remains notoriously sample inefficient. 
A key factor to unlock its potential has been the advent of massive parallel computation, which speeds up experience collection by running independent copies of an environment simultaneously. Training RL on parallel processes has become standard nowadays.While this practice reduces wall-clock time by a factor proportional to the number of parallel processes, it does not fully exploit parallelization at the algorithmic level, which mostly follows the same design of a sequential computation setting~\citep{Impala, espeholt2019seed}. Critically, little has been done to address the interplay between parallel computation and the root cause of RL's inefficiency, i.e., exploration.

Recent works have provided some compelling ideas in this direction. \citep{mayor2025impact} consider the trade-off between the length of each rollout taken from the environment with the number of parallel environments from which rollout are taken simultaneously. Their findings show that parallel environments make for a better use of computational resources than increasing the rollout length, despite the total number of collected samples being equal. This is made possible by avoiding resets in the parallel environments, which provide more varied exploration than sampling longer rollout from a fixed starting state distribution.

On an orthogonal line, \citet{paola2025enhancing} consider reward-free exploration in the parallel setting. While exploration is usually balanced with exploitation in the standard RL setting, in a reward-free one there is no reward to exploit. In this scenario, a compelling goal for exploration is to collect the most useful experience to be able to solve any RL task downstream.
A common and principled formulation for this goal is state entropy maximization \citep{hazan2019provably}, in which the agent aims to maximize the entropy of the distribution of state visitations induced by its policy. This has been shown to encourage the discovery of varied behaviors in an unsupervised manner, and to provide a useful initialization for RL~\citep{mutti2021task, liu2021behavior, seo2021state, yarats2021reinforcement}. However, most of these approaches consider a single generalist exploration policy. 

In a departure from those previous works, \cite{paola2025enhancing} aim at learning a set of policies that \emph{collectively} maximize the state distribution entropy, while preserving specialization within the set. While they provide both theoretical fundamentals and promising results, their empirical investigation is limited to toy domains with discrete states and action spaces and a few parallel processes.In this paper, we extend the problem of parallel state entropy maximization to continuous domains and massive parallelization. The main contribution of our work is to develop a practical and scalable algorithm, called \textbf{K-Myriad}, which is able to pretrain a huge collection of policies (up to 50) while interacting with hundreds or thousands copies of the same environment.

Moreover, we show how the collection of policy can be employed to improve the efficiency of downstream RL for a variety of tasks. With respect to a single general-purpose maximum entropy policy, the collection of policies has an higher chance of including a policy that is well aligned with the task reward, whatever the task. Through a straightforward evaluation of the most promising policy in the collection for a given task, which is then used as the initial policy of a base RL algorithm, our approach is able to outperform the same base algorithm starting from a random initialization or a generalist maximum entropy policy.

The \textbf{contributions} are organized as follows:
\begin{itemize}[topsep=0pt, itemsep=0pt, leftmargin=10pt]
    \item We define the problem of state entropy maximization in large-scale parallel MDPs (Section~\ref{sec:problem});
    \item We propose a scalable policy gradient algorithm, K-Myriad, to address the introduced problem in challenging high-dimensional domains and hundreds of parallel processes (Section~\ref{sec:exploration});
    \item We propose a simple approach to jump start RL with the trained parallel policies that is corroborated by an illustrative experiment in a continuous low-dimensional domain (Section~\ref{sec:jumpstarting});
    \item We present an experimental campaign on GPU-based parallel simulators of high-dimensional continuous control tasks, mostly variations of the Ant domain of Isaac Sim (Section~\ref{sec:experiments}).
\end{itemize}

The complete codebase of our work is available \href{https://anonymous.4open.science/r/K-Myriad-C46C/README.md}{\underline{here}}.

\section{Related Works}
\label{sec:relatedworks}

\textbf{State Entropy Maximization.}~~
Maximizing the entropy of state visitation in Markov decision processes was first introduced by \citet{hazan2019provably}, inspiring a broad range of follow-up work on reward-free and task-agnostic exploration 
\citep{lee2019smm,mutti2020intrinsically,mutti2021task,mutti2022importance,mutti2022unsupervised,mutti2023unsupervised,guo2021geometric,liu2021behavior,liu2021aps,seo2021state,yarats2021reinforcement,nedergaard2022k,yang2023cem,tiapkin2023fast,jain2023maximum,kim2023accelerating,zisselman2023explore,zamboni2024limits,zamboni2024explore,zamboni2025towards,paola2025enhancing}.  
These works show the effectiveness of entropy maximization for policy pretraining~\citep[e.g.,]{liu2021behavior}, as an intrinsic objective for exploration~\citep[e.g.,][]{seo2021state}, and as a regularization to improve robustness of the learned policy~\citep{ashlag2025state}.

\textbf{Exploration with Multiple Agents.}~~
Concurrent exploration with multiple agents has been studied in settings where agents operate in the same environment~\citep{guo2015concurrent,parisotto2019concurrent,jung2020population,chen2022society,qin2022analysis}. 
Other works investigate information sharing between agents \citep{alfredo2017efficient,holen2023loss}, or the theoretical foundations of coordinated exploration \citep{dimakopoulou2018coordinated,dimakopoulou2018scalable}.  
In multi-agent reinforcement learning~\citep[MARL,][]{albrech2024multiagent}, exploration has often been driven by reward shaping and heuristic bonuses \citep{wang2019influence,zhang2021made,xu2024population}.  
For instance, \citet{zhang2023self} propose an exploration bonus that maximizes deviation from jointly explored regions. \citep{zamboni2025towards} explicitly address state entropy maximization in a multi-agent setting. 
A key distinction is that, in conventional MARL, agents act concurrently in a shared environment, and their trajectories are interdependent \citep{DBLP:journals/corr/abs-2106-02195,DBLP:journals/corr/abs-2112-11701,pmlr-v139-lupu21a}.  
By contrast, our framework involves agents interacting with independent replicas of the environment. This structural difference eliminates inter-agent interference while allowing specialization to emerge, setting our work aside from the MARL literature.

\textbf{Efficient Methods for Parallel Learning.}~~
A related line of work focuses on systems and architectures that scale RL via high-throughput parallel data collection and accelerator-aware training. 
IMPALA~\citep{Impala} decouples acting and learning with a centralized learner and many distributed actors, using V-trace for off-policy correction to enable stable learning at very high frame rates. 
SEED RL further optimizes the pipeline through centralized, accelerator-backed inference and an efficient communication layer, attaining millions of frames per second at reduced cost~\citep{espeholt2019seed}. 
On single machines, Sample Factory demonstrates aggressive CPU–GPU pipelining and asynchronous PPO variants that reach $>10^5$ FPS~\citep{petrenko2020samplefactory}. 
For multi-agent settings, WarpDrive executes both simulation and learning entirely on the GPU, eliminating CPU--GPU data transfers and scaling to thousands of concurrent environments and agents~\citep{lan2022warpdrive}. 
These systems are complementary to our PMDP formulation: While our objective targets diversity and entropy maximization across parallel environment replicas, the above frameworks provide engineering patterns (centralized learners, accelerator-based inference, end-to-end GPU execution) that can be used to implement our algorithm efficiently at scale. 
In robotics, parallel simulators have also been shown to drastically shorten wall-clock time~\citep[e.g., Isaac-based setups,][]{rudin2022learning,mittal2023orbit}. Also related to our setting, but mostly orthogonal in spirit, the work by~\citet{mayor2025impact} analyzes the impact of parallel data collection for on-policy reinforcement learning, where the same policy is deployed on the parallel environments.

\textbf{Diverse Skill Discovery.}~~
The problem of learning a diverse set of skills is closely related to diversifying exploration through parallelization. 
Notable approaches include variational intrinsic control~\citep{gregor2017variational}, DIAYN~\citep{eysenbach2018diversity}, and their extensions that maximize mutual information between skills and visited states \citep{hansen2019fast,sharma2020dynamics,campos2020explore,liu2021aps,he2022wasserstein}.  
\citet{zahavy2022discovering} explicitly optimize for diversity among policies.  
These methods complement ours through alternative heuristics to encourage behavioral variety among agents, without relying on extrinsic rewards. Unlike ours, they do not consider the interplay between agents' diversity and parallel computation.

\section{Preliminaries}
\label{sec:preliminaries}


Below we introduce the main tools and definitions we will make use of throughout the paper.

\textbf{Notation.}~~
We denote $[N] := \{1,2,\dots,N\}$. For a space $\mathcal{X}$, $\mathcal{X}^m$ is the product space of $m$ copies of $\mathcal{X}$.

\begin{figure}[t] 
  \centering
\includegraphics[width=0.75\linewidth]{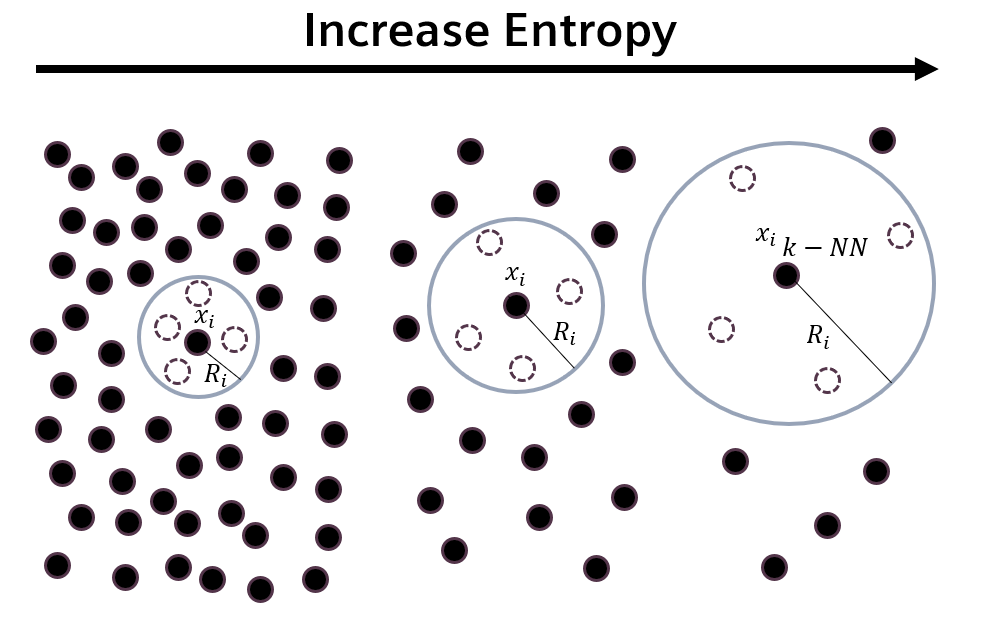}
  \caption{Visualization of the k-NN entropy estimator (Eq.~\ref{eq:maxentropyestimator}).
  }
  \label{fig:preliminaries-overview}
\end{figure}

\textbf{Entropy.}~~
For a measurable space $\mathcal{X}$, we denote the set of all probability density functions (PDFs) on $\mathcal{X}$ as $\mathcal{P}(\mathcal{X})$.
For a random variable $X$ on $\mathcal{X}$ with PDF $f \in \mathcal{P}(\mathcal{X})$, we denote its differential Shannon entropy as $H(X) = - \int_{\mathcal{X}} f(x) \log f(x) \, dx.$ In continuous, high-dimensional settings, computing the latter integral can be non-trivial, as it requires to characterize the PDF $f$.
Nonetheless, various methods exist to estimate the entropy $H(X)$ given realizations $\{ x_i \}_{i = 1}^N$ of the random variable $X$ \citep{Beirlant1997NonparametricEE}.
A common approach \citep{Singh01022003} is to use a $k$-Nearest Neighbors (k-NN) estimator of the form 
\begin{equation}
\label{eq:maxentropyestimator}
\hat{H}_k(f) = -\frac{1}{N} \sum_{i=1}^N \ln \frac{k}{N V_i^k} + \ln k - \Psi(k),  
\end{equation}
where $\Psi$ is the digamma, $\ln k - \Psi(k)$ is a bias correction term, and $V_i^k$ is the volume of the hyper-sphere of radius 
$R_i = \|x_i - x_i^{k\text{-NN}}\|, $ which is the Euclidean distance between $x_i$ and its k-NN $x_i^{k\text{-NN}}$. The k-NN estimator is known to be asymptotically unbiased and consistent~\citep{Singh01022003}. Fig.~\ref{fig:preliminaries-overview} provides a visualization of how the local density around a point $x_i$ is related to distance $R_i$ from their k-NN.


\textbf{MDP.}~~
The Markov Decision Process~\citep[MDP,][]{puterman2014markov} is a popular model for sequential decision making under uncertainty. The MDP is defined as $\mathbb{M} := (\mathcal{S}, \mathcal{A}, p, r, \mu, T)$ where $\mathcal{S} \subseteq \mathbb{R}^{d_s}$ is a continuous state space, 
$\mathcal{A} \subseteq \mathbb{R}^{d_a}$ is a continuous action space, $p$ is a transition kernel such that $p(\cdot \mid s,a) \in \mathcal{P}(\mathcal{S})$ is the PDF of the next state given the current state and action pair $(s,a) \in \mathcal{S} \times \mathcal{A}$, $r$ is a reward function such that $r (s, a) \in \mathbb{R}$ is the payoff for  $(s,a) \in \mathcal{S} \times \mathcal{A}$,  $\mu \in \mathcal{P}(\mathcal{S})$ is the initial state PDF, and $T \in \mathbb{N}^+$ is the finite horizon of an episode of interaction.

\textbf{Policy and state distribution.}~~
An agent can exercise partial control over the MDP evolution through their actions. A policy $\pi : \mathcal{S} \to \mathcal{P} (\mathcal{A})$ describe the strategy to which actions are selected such that $\pi (a | s)$ is the conditional PDF of $a \in \mathcal{A}$ in $s \in \mathcal{S}$. Rolling out a policy over the MDP induces a distribution over trajectories $\tau = (s_0, a_0, \ldots, s_{T - 1}, a_{T - 1}, s_T) \in \mathcal{K}$ with PDF $p_\pi (\tau) = \mu(s_0) \prod_{t = 0}^{T - 1} \pi(a_t | s_t) p (s_{t + 1} | a_t, s_t)$ and a distribution over states $d_\pi \in \mathcal{P} (\mathcal{S})$ with PDF $d_\pi (s) = \mathbb{E}_{\tau \sim p_\pi} [\sum_{t = 0}^{T - 1} \mathds{1} (s_t = s | \tau) / T]$, which captures the probability of visiting a state $s$, marginalized over the steps of a trajectory of length $T$.

\textbf{Objective.}~~
The typical objective of an MDP is to maximize the expected cumulative sum of rewards collected within a trajectory. Formally, $\max_{\pi \in \Pi} \mathbb{E}_{\tau \sim p^\pi} [\sum_{t = 0}^{T - 1} r (s_t, a_t)]$ where $\Pi$ denotes the space of Markovian policies and $\pi^*$ denotes the policy attaining the maximum, hence the \emph{optimal policy}. The objective can be equivalently written as $\max_{\pi \in \Pi} \int_{\mathcal{S} \times \mathcal{A}} d_\pi (s) \pi (a | s) r (s, a) ds$, which highlights that the problem is linear in the state distribution density $d_\pi$. Reinforcement Learning~\citep[RL,][]{sutton} is a family of methods to \emph{learn} the optimal policy from sampled interactions with MDP. RL methods typically employ the concept of value functions, which are $V^\pi_t (s) = \mathbb{E}_{\tau \sim p^\pi} [\sum_{j = t}^{T - 1} r(s_j, a_j) | s_t = s]$ and $Q^\pi_t (s, a) = \mathbb{E}_{\tau \sim p^\pi} [\sum_{j = t}^{T - 1} r(s_j, a_j) | s_t = s, a_t = a]$.


\section{Unsupervised Exploration in Parallel MDPs}
\label{sec:problem}

\begin{figure}[t] 
  \centering
  \includegraphics[width=0.75\linewidth]{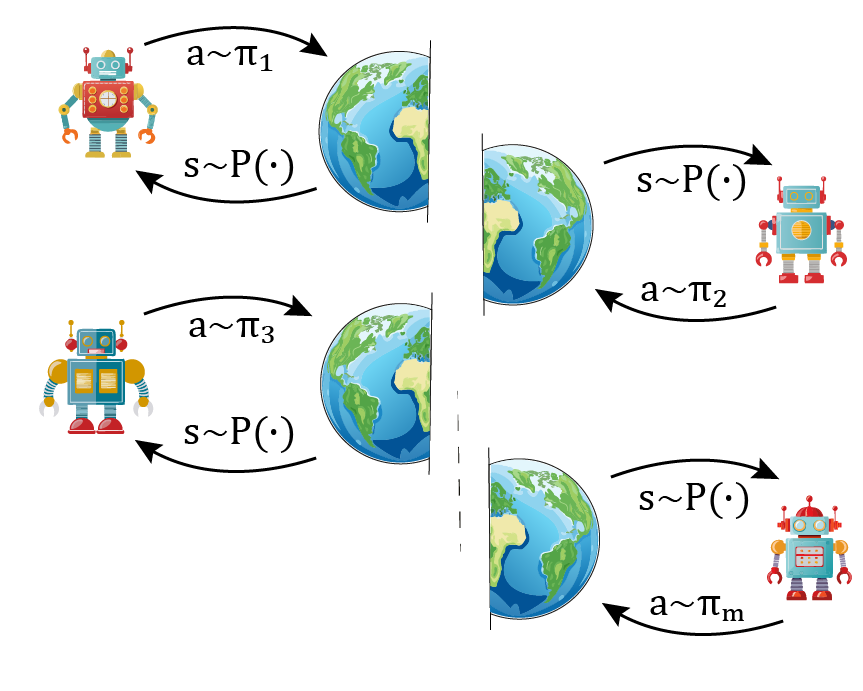}
  \vspace{-15pt}
  \caption{The parallel agent-environment interaction.}
  \label{fig:preliminaries-pmdp}
\end{figure}

We are interested in a setting in which computational infrastructure allows for simulating a multitude of copies of the same environment in parallel, nowadays the most common scenario for RL in practice. We formalize this setting as a Parallel Markov Decision Process~\citep[PMDP,][]{sucar2007parallel}, i.e., a collection 
$$\mathbb{M}_p = (\mathbb{M}_i)_{i \in [m]} \qquad \text{ where } \mathbb{M}_i = (\mathcal{S}, \mathcal{A}, p, \mu, T)$$
are $m \in \mathbb{N}^+$ identical and independent copies of the MDP $\mathbb{M}_i$.\footnote{The MDP definition does not include a reward function. For reasons that will be clear later, we consider a reward-free setting.}
The agent interacts with the PMDP by means of a \emph{parallel policy}, which is a collection $\pi_p = (\pi_i)_{i \in [m]}$ of, possibly different, stochastic Markovian policies $\pi_i : \mathcal{S} \to \mathcal{P} (\mathcal{A})$, one for each copy of the MDP $\mathbb{M}_i$ (see Fig.~\ref{fig:preliminaries-pmdp}).

The interaction protocol of an agent with the PMDP proceeds as follows. 
At the beginning of the episode, the agent sends a policy $\pi_i$ to the $m$ parallel processes simulating the PMDP.
For each process $i \in [m]$, an initial state is drawn $s^i_0 \sim \mu$ independently. Then, for each step $t \in \{0, \ldots, T-1\}$, an action $a^i_t \sim \pi_i (\cdot \mid s_t^i)$ is drawn from the respective policies independently, so that they trigger a transition in their corresponding copies of the MDP as $s_{t + 1}^i \sim P(\cdot \mid s_t^i, a_t^i)$. Each parallel episode ends when the final state $s_T^i$ is reached. Hence, the processes send the collected trajectories $\tau_i = (s_0^i, a_0^i, \dots, s_{T-1}^i, a_{T-1}^i, s_T^i)$ back to the agent, which gathers them into a \emph{parallel trajectory} $\tau := (\tau_i)_{i \in [m]} \in \big(\mathcal{S}^{T+1} \times \mathcal{A}^T\big)^m$ stored with the previous experience, serving the basis for updating the policies to be sent to the processes in the subsequent episode.

\subsection{The Parallel State Entropy Objective}

The agent typically updates the parallel policy in the direction of the gradient of the RL objective, i.e., the maximization of the expected return. Here we consider a reward-free setting, in which the agent can gather experience from the PMDP to pre-train a parallel policy, before being tasked with a reward function in a variety of downstream tasks. To this end, following a recent problem formulation by~\citet{paola2025enhancing}, we aim to learn a collection of policy that is (a) \emph{diverse}, to provide a robust initialization to downstream tasks and to implicitly drive RL to find alternative task solutions, (b)  \emph{exploratory}, such that the policies collectively cover the states of the environment, which allows to identify rewarding region and to learn from there.

Formally, \citet{paola2025enhancing} generalize the maximum state entropy objective~\citep{hazan2019provably} to PMDPs. Let $d_{\pi_i} \in \mathcal{P}(\mathcal{S})$ the state distribution induced by the $i$-th component of the parallel policy $\pi_p$ on the respective process $i$, i.e., $d_{\pi_i} (s) = \mathbb{E}_{\tau_i \sim p_{\pi_i}} [\sum_{t = 0}^{T - 1} \mathds{1} (s_t = s | \tau_i) / T]$. Then, we formulate the \emph{parallel} maximum state entropy objective as
\begin{align}\label{eq:mse_parallel}
    \max_{\pi_p \in \Pi^m} &\mathcal{J} (\pi_p)  = \max_{\pi_p \in \Pi^m} \left\{   H (d_{\pi_p}) \right\} \\ &\text{ where } d_{\pi_p}(s) = \frac{1}{m} \sum_{i=1}^m d_{\pi_i}(s) \nonumber
\end{align}
is the \emph{parallel state distribution}, the cumulative state distribution induced by the parallel policy $\pi_p$ across the PDMP. How can~\ref{eq:mse_parallel} be optimized in practice, by relying on sampled interactions with the environment? \citet{paola2025enhancing} proposed a flexible policy gradient~\citep{williams1992simple, sutton1999policy, peters2008reinforcement} scheme that can be summarized in the following steps, repeated until convergence.
\begin{itemize}[topsep=0pt, itemsep=0pt, leftmargin=10pt]
    \item \textbf{Collecting rollouts}: The parallel policy run each component $\pi_i$ in their respective environment copy to collect a parallel trajectory.
    \item \textbf{Entropy estimation}: The parallel trajectory is used to estimate the parallel state distribution $d_{\pi_p}$ to then compute its entropy, or to estimate the entropy as $H(d_{\pi_p})$ directly.
    \item \textbf{Policy update}: Each component of $\pi_p$ is updated in the direction of their respective gradient  $$\pi'_i \gets \pi_i + \alpha \nabla_{\pi_i} H (d_{\pi_p}) \qquad \forall i \in [m]$$
\end{itemize}
While \cite{paola2025enhancing} have shown this recipe to work, they limited their analysis to toy domains with discrete states and actions and when considering a handful of parallel agents at most. It is less understood how to scale policy gradient algorithms for state entropy maximization on more challenging domains, with high-dimensional continuous states and action spaces, and massive parallelization, when hundreds or thousands of policies are deployed in parallel.

\section{Scalable Parallel Algorithms for the State Entropy Objective}
\label{sec:exploration}

The primary contribution of our work is a \emph{scalable} algorithm, called \emph{K-Myriad}, to learn a population of policies that collectively maximize the state entropy objective defined in~\ref{eq:mse_parallel}, facing continuous, high-dimensional domains and massively parallelized environments.
The algorithm roughly follows the steps of the policy gradient scheme presented above, with an implementation guided by scalability.

\subsection{Policy Architecture}
The first aspect to be addressed in designing a scalable algorithm is how to represent the parallel policy. In principle, we would like to send a separate policy network to each parallel process. In practice, parallel processes are often run on the same GPU with shared memory, making it intractable to store dozens or hundreds of fully independent parameterizations. 

\begin{figure}
    \centering
    \includegraphics[width=\linewidth]{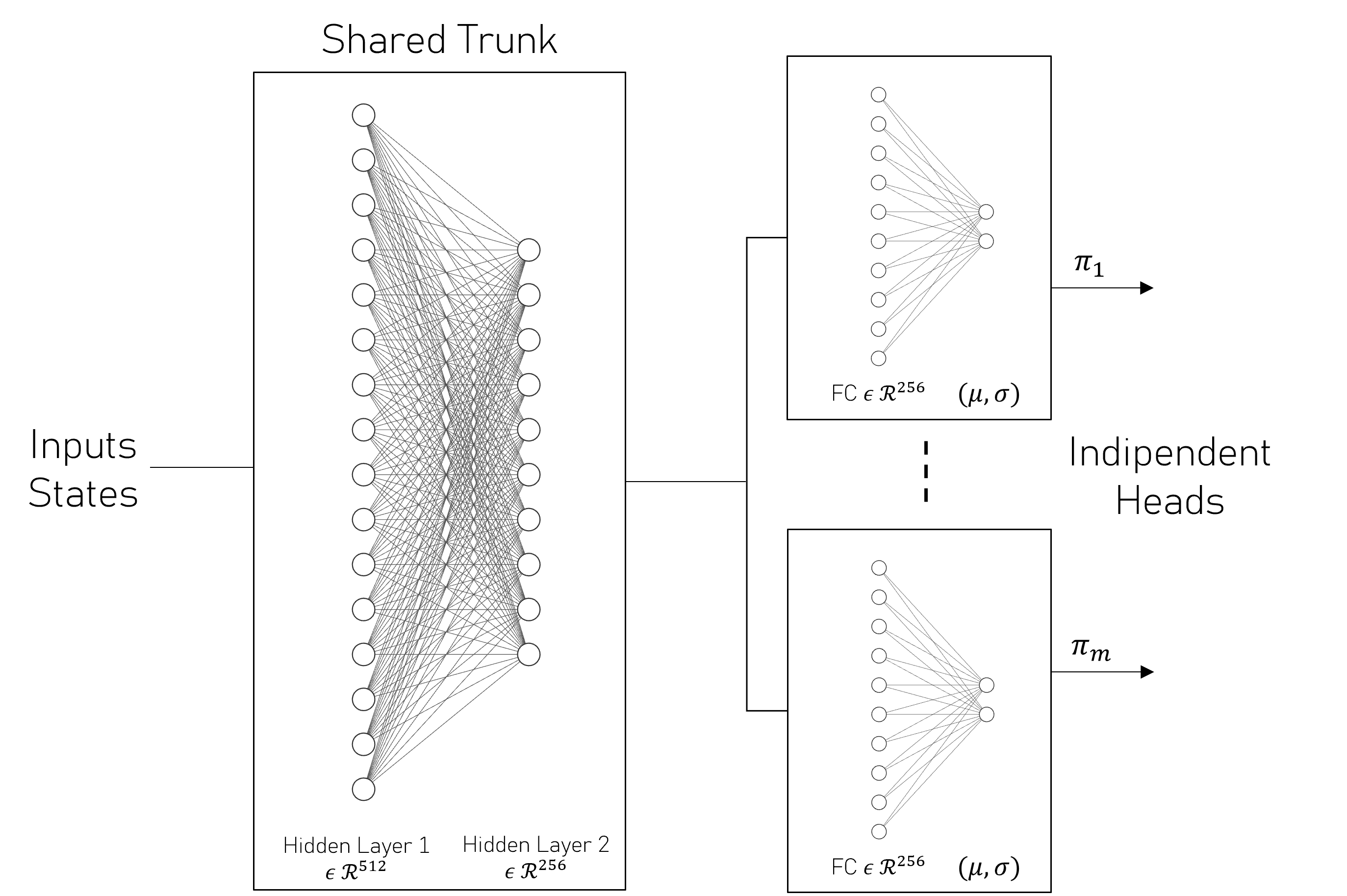}
    \caption{K-Myriad policy architecture. 
A shared backbone extracts features from the input, while independent heads produce distinct policy outputs.} 
    \label{fig:sharednetwork}
\end{figure}

In K-Myriad, policies are implemented within a single policy network, composed of a shared trunk and $m$ independent heads, one for each deployed policy (see Fig.~\ref{fig:sharednetwork}). The shared trunk has two fully connected layers that process the state variables into a representation to be sent to the independent heads. The heads are shallow networks that take the representation and produce the mean and standard deviation of a Gaussian action distribution. This architectural design has two significant benefits: First, it allows for saving memory as we need to store the parameters of the shared trunk once for all the processes; Second, it allows to train a shared representation---as we shall see, the shared trunk collects gradients from all the independent processes---while keeping specialized action selection strategies for the parallel copies. The single network architecture also allows for sampling actions across all parallel copies with a single forward pass, substantially speeding up computation.

\subsection{Loss Function}
Having defined the policy architecture, we now need to specify a loss function to train the parameters of the shared trunk and the independent heads, thereby maximizing the state entropy objective~\ref{eq:mse_parallel}. As it is common in prior works~\citep[e.g.,][]{mutti2021task, liu2021behavior}, we turn to a direct estimation of the entropy $H(d_{\pi_p})$ without an explicit estimate of the state distribution $d_{\pi_p}$, which is problematic in high-dimensional domains. We can use the kNN entropy estimator~\citep{Singh01022003} we presented in~\ref{eq:maxentropyestimator}. 
For a parallel trajectory $\tau = ( (s_0^i, a_0^i, \dots, s_{T-1}^i, a_{T-1}^i, s_T^i)_{i \in [m]} )$, we estimate the entropy as follows
\begin{equation}
\label{eq:loss}
\hat{H}_k \propto \sum_{i=1}^{m} \sum_{t = 0}^{T} \ln  \|s_t^i - \text{kNN}(s_t^i) \|^d,  
\end{equation}
where $\text{kNN}(s_t^i)$ is the kNN of the state $s_t^i$.\footnote{We omit all the constants, which are the same as in~\ref{eq:maxentropyestimator}.}
Since the states are sampled by rolling out the parallel policy, they preserve a dependence on the policy parameters and, especially, the state $s_t^i$ depends on the parameters of the head $i$ and the shared trunk. Thus, we can differentiate the equation~\ref{eq:loss} with respect to the policy parameters to perform a step in the gradient direction. For the details of the loss implementation please see the dedicated Appendix~\ref{sec:LossCalculation}. It is crucial to note that the $\text{kNN}(s_t^i)$ may be a state visited in a copy of the environment $j \neq i$, so that the independent heads have an incentive to specialize on different regions of the state space.

\subsection{The Algorithm}
We now have the elements to present K-Myriad, whose pseudocode is reported in Algorithm~\ref{alg:kmyriad}. First, a policy is initialized according to the architecture described in the previous sections. 
In one epoch of the main loop, each parallel process is assigned to a policy head. In the most straightforward case, the mapping is one-to-one, with a policy having $m$ independent heads, as many as the parallel processes. In general, however, one can decide to run the same head on more than one process. This will be the case in our experiments, in which we will consider up to 1000 parallel processes for 10 or 50 heads. In this way, we can both reduce the total number of parameters and collect multiple trajectories from the same policy in a single simultaneous rollout.

\begin{figure}
\begin{minipage}[ht]{\linewidth}
\begin{tcolorbox}[colback=softbluegray!10, colframe=softbluegray!60, boxrule=0.5pt, arc=4pt, width=\linewidth, coltext=black, coltitle=black, title=\textbf{Algorithm 1}: K-Myriad ]
\refstepcounter{algorithm}
\label{alg:kmyriad}
\begin{algorithmic}
    \STATE \textbf{Input}: Processes $m$, Policies $\overline{m}$, Rate $\alpha$
    \STATE Initialize the policy $\pi_\theta$
    \FOR{epoch $e = 1, \dots E$}
        \STATE \textbf{Policies to Processes:}
        \STATE Assign a policy head to each process
        \STATE \textbf{Data Collection:}
        \STATE Collect a parallel trajectory $\tau$ as follows
        \STATE \textbf{Entropy Estimation:}
        \STATE Compute $\hat H_k$ as in Eq.~\ref{eq:loss} given $\tau$
        \STATE \textbf{Policy Gradient Update:}
        \STATE $L(\theta) = \hat H_k$
        \STATE $\theta \gets \theta + \alpha \nabla_\theta L(\theta)$
    \ENDFOR
    \STATE \textbf{Output}: Parallel policy $\pi_\theta$
\end{algorithmic}
\end{tcolorbox}
\end{minipage}
\end{figure}

After the assignment, a rollout is collected in parallel for all processes, along with their respective policy heads. With the collected data, the loss is computed through the entropy estimation presented previously. Finally, the gradient of the loss is propagated in a single backward pass to the entire policy network, shared trunk, and independent heads. Once the epoch exceeds the budget $E$, the algorithm ends and the parallel policy is given as output.

\section{Jump Starting RL with a Pretrained Parallel Policy}
\label{sec:jumpstarting}

The parallel policy trained with K-Myriad provides a diverse set of exploratory behaviors. In this section, we describe how these policies can be used to \emph{jump start} RL of downstream tasks, described by a reward function in the same environment. The most straightforward solution is to initialize the policy network of any RL algorithm of choice with the pretrained policy~\citep{mutti2021task, liu2021behavior}. However, the parallel setting opens a wider range of possibilities: Which policy of the collection do we deploy in which copy of the environment? In this paper, we propose a simple yet effective strategy for evaluating policies that are more \emph{aligned} with a given task and learning from them. While a more sophisticated approach could be considered, our aim is to showcase the potential of parallel pretraining rather than providing a definitive answer to the jump-start RL problem. Before going to more challenging experiments, we provide helpful intuition with an illustrative example.

\begin{figure}[ht]
    \centering
    \begin{subfigure}{\linewidth}
        \centering
        \includegraphics[width=0.28\linewidth]{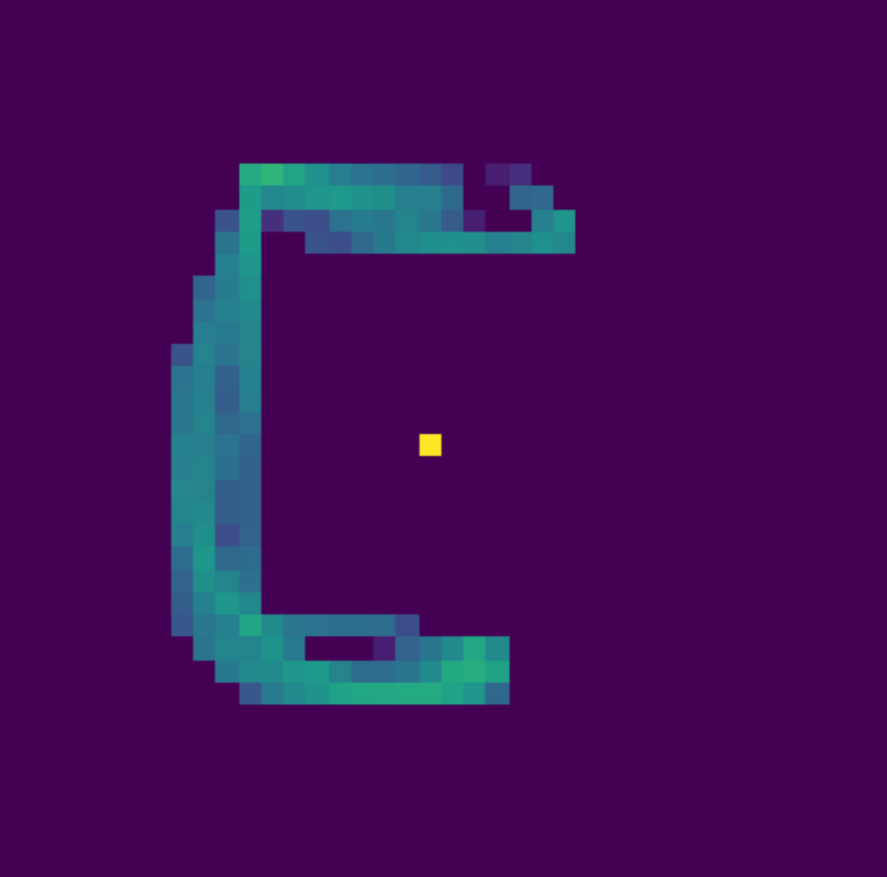}
        \includegraphics[width=0.28\linewidth]{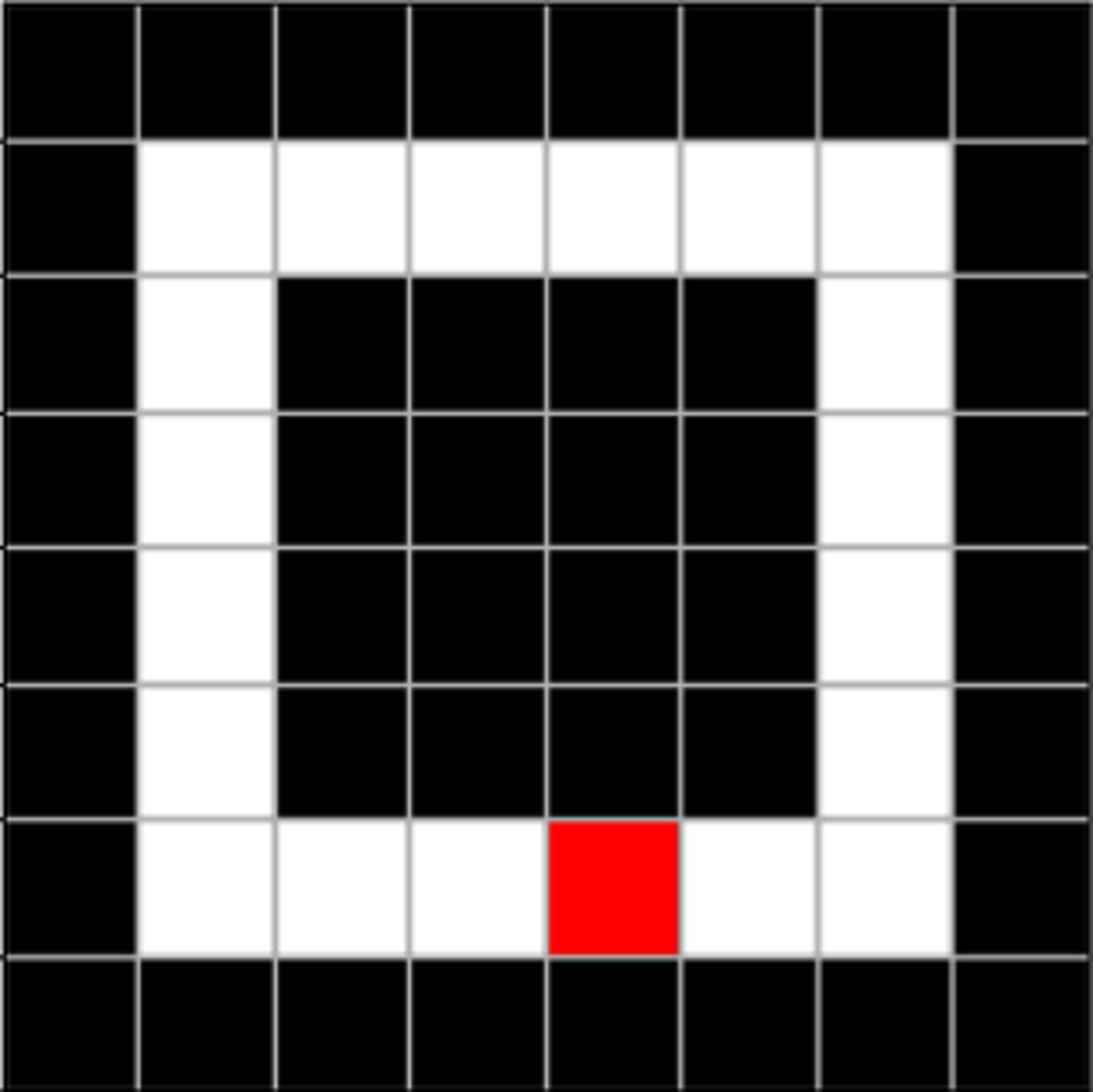}
        \includegraphics[width=0.28\linewidth]{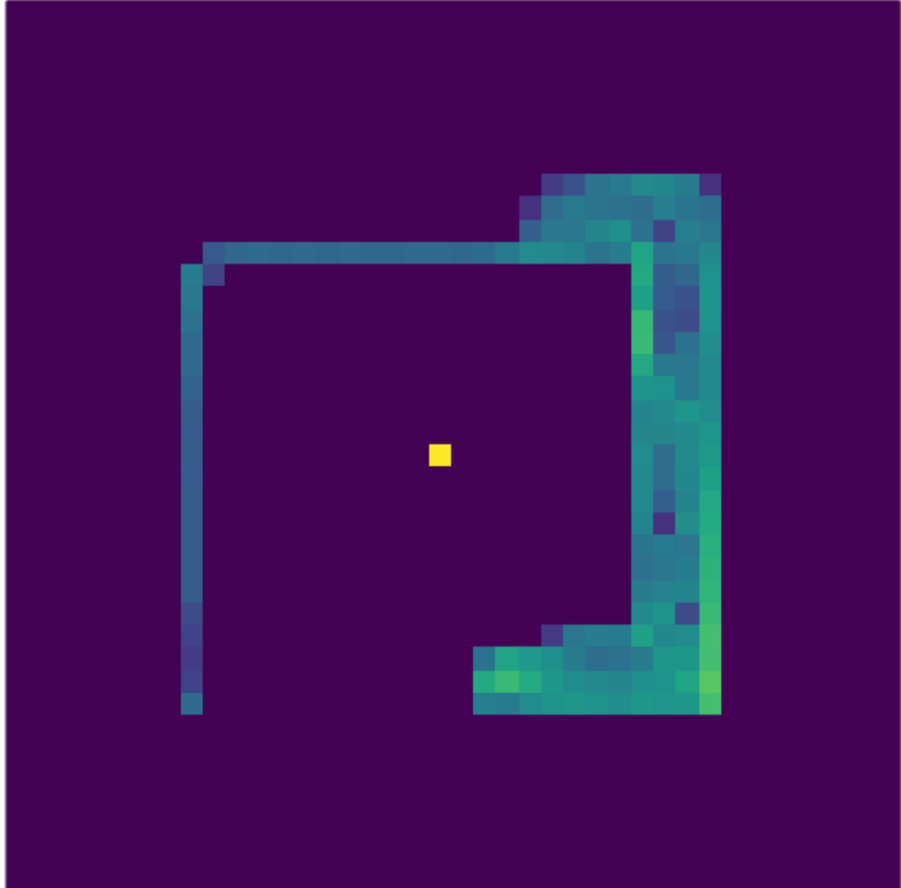}
        \vspace{-2pt}
        \caption{Pretraining}
    \end{subfigure}

    \begin{subfigure}{\linewidth}
        \centering
        \includegraphics[width=0.28\linewidth]{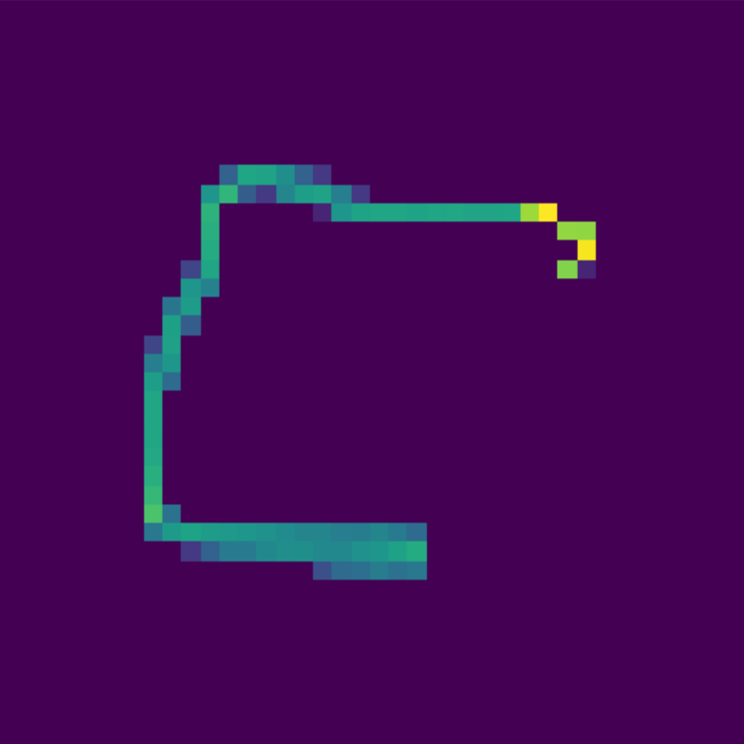}
        \includegraphics[width=0.28\linewidth]{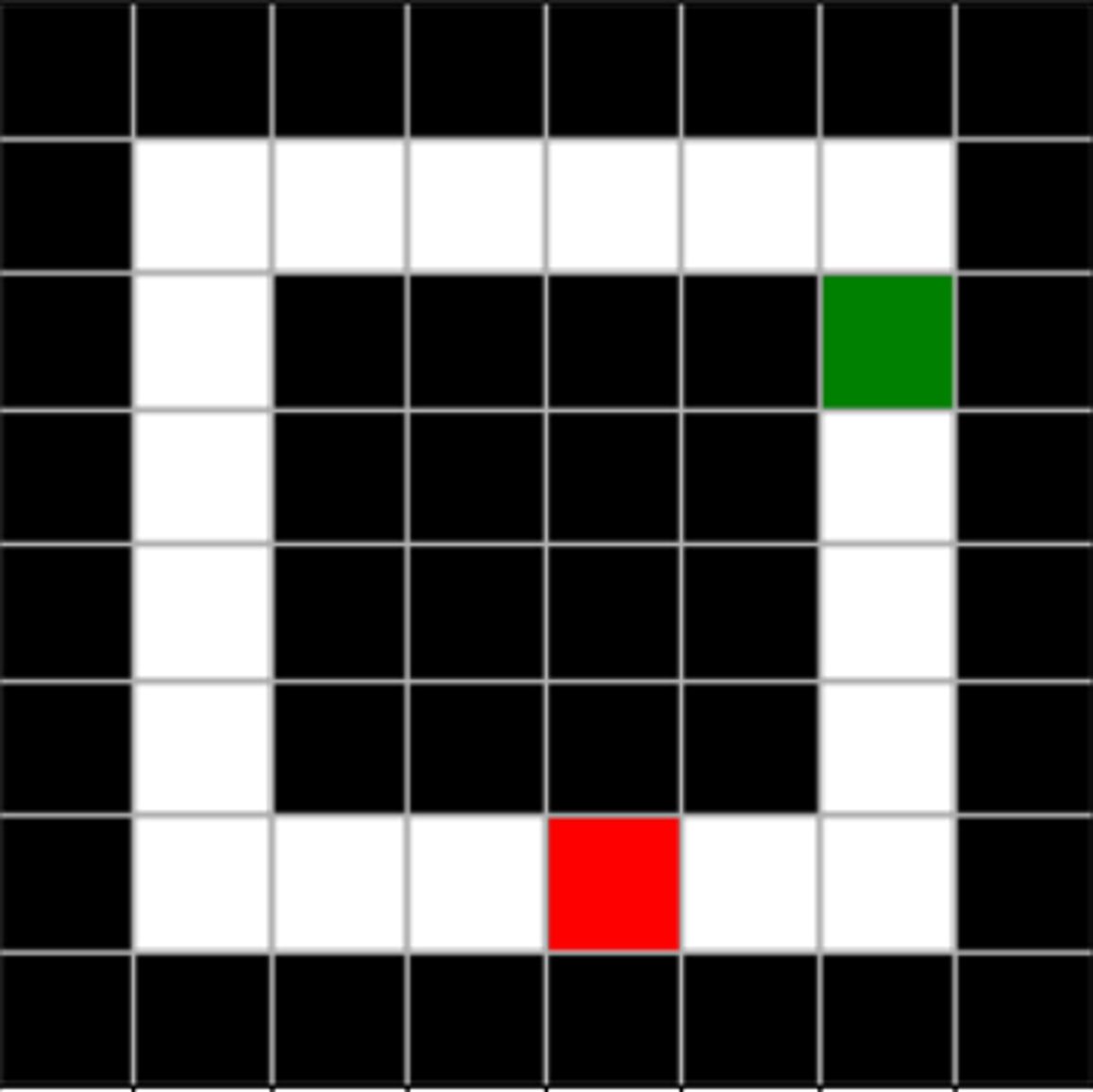}
        \includegraphics[width=0.28\linewidth]{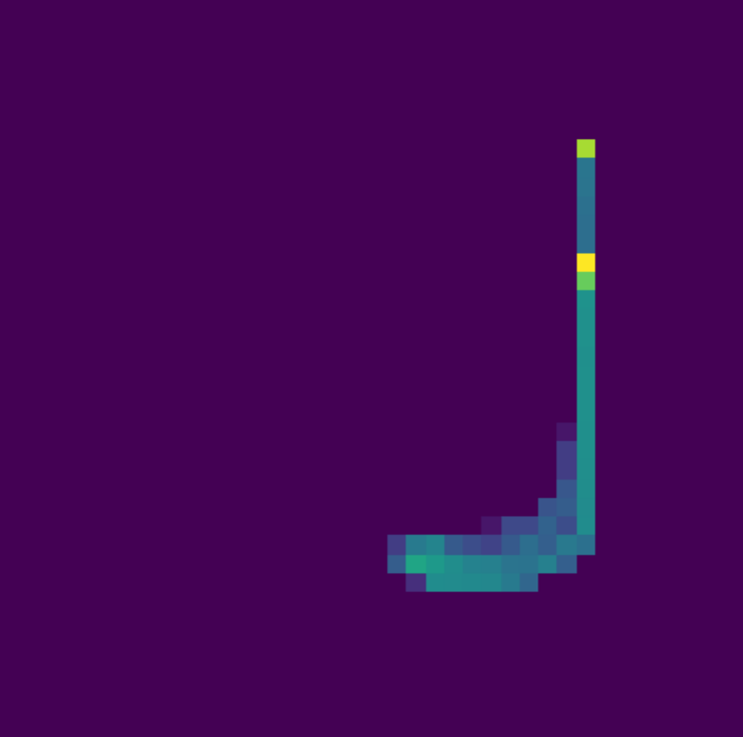}
        \vspace{-2pt}
        \caption{Jump start with top right goal}
    \end{subfigure}

    \begin{subfigure}{\linewidth}
        \centering
        \includegraphics[width=0.28\linewidth]{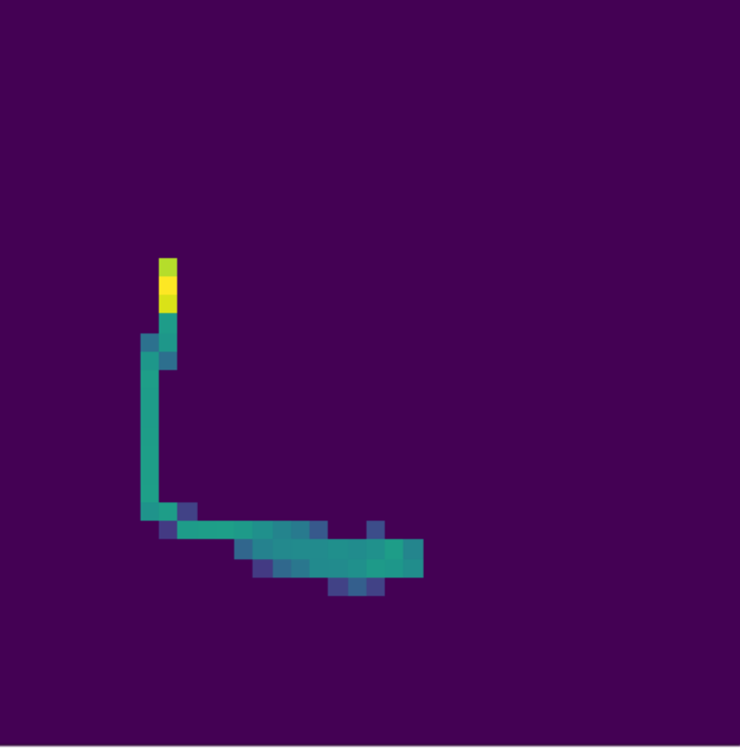}
        \includegraphics[width=0.28\linewidth]{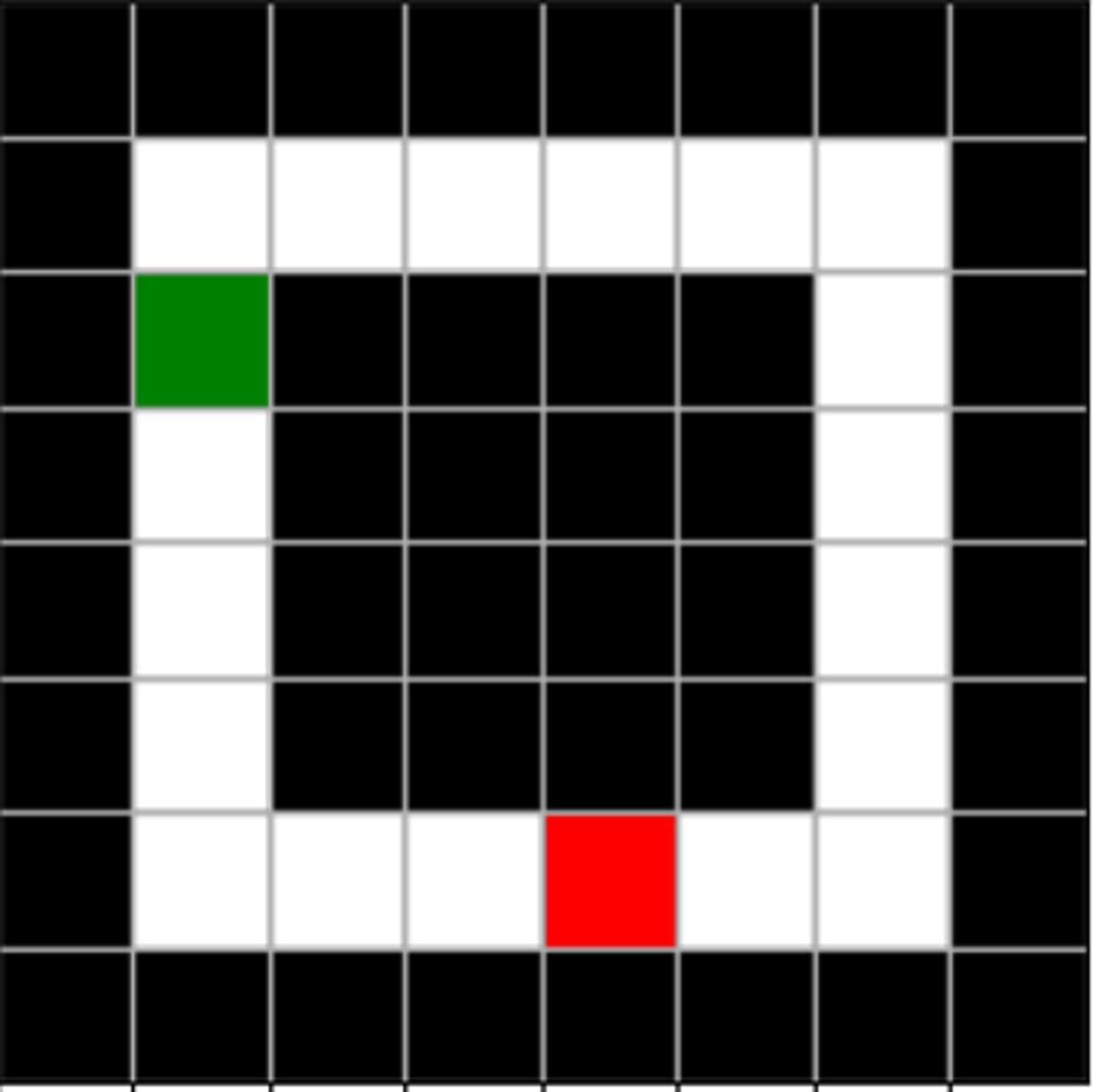}
        \includegraphics[width=0.28\linewidth]{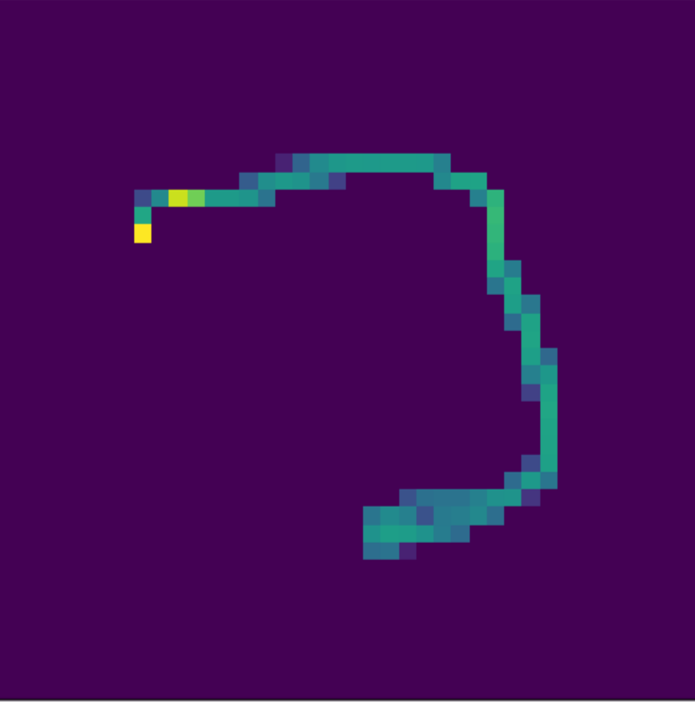}
        \vspace{-2pt}
        \caption{Jump start with top left goal}
    \end{subfigure}
    
    \vspace{-5pt}
    \caption{The \texttt{PointMaze} environment. (a) Pretraining with K-Myriad and resulting heatmaps of two parallel agents. (b,c) Jump-starting RL for different goal states with the two pretrained policies (heatmaps after training). Goal cells are in green, and the initial state is in red.}
    \label{fig:stateentropy-examples}
\end{figure}

\subsection{Illustrative Example}

In this setup, we consider the \emph{PointMaze} environment from \texttt{gym-robotics}, a low-dimensional yet continuous domain. Here, two agents are first trained to maximize the parallel state entropy (Eq.~\ref{eq:mse_parallel}). As we can see from Fig.~\ref{fig:stateentropy-examples} (a), they produce behaviors that are heterogeneous while covering the state space. When a downstream task is given (b,c), we can initialize any RL algorithm with the pre-trained policies. It is interesting to observe that the different initializations result in distinct strategies for solving the task, for example, reaching a goal state at the top right of the environment by traversing the maze in opposite directions.

\section{Experiments}
\label{sec:experiments}

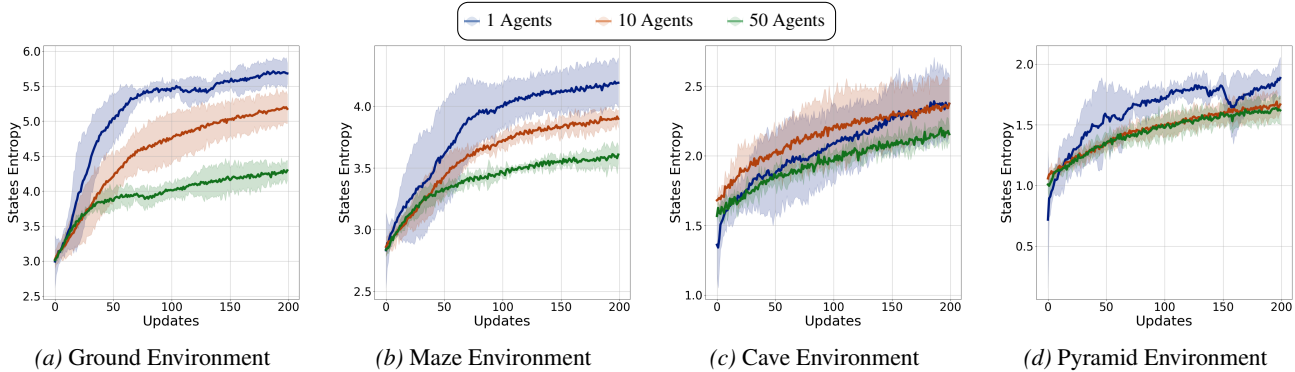
\begin{figure*}[ht!] 
    \centering
    \begin{tikzpicture}
    \node[draw=black, rounded corners, inner sep=2pt, fill=white] (legend) at (0,0) {
        \begin{tikzpicture}[scale=0.7]
            \def\linelen{0.3}
            \def\linegap{0.05}
            \def\textgap{0.01}
            \def\y{0}

            \def\xA{0}
            \def\xB{2.5}
            \def\xC{5.0}

            \draw[thick, color={rgb,255:red,76; green,114; blue,176}, opacity=0.8] (\xA,\y) -- ({\xA + \linelen},\y);
            \fill[color={rgb,255:red,76; green,114; blue,176}, opacity=0.2] (\xA,{\y - 0.1}) rectangle ({\xA + \linelen},{\y + 0.1});
            \node[anchor=west, font=\scriptsize] at ({\xA + \linelen + \textgap}, \y) {1 Agents};

            \draw[thick, color={rgb,255:red,221; green,132; blue,82}, opacity=0.8] (\xB,\y) -- ({\xB + \linelen},\y);
            \fill[color={rgb,255:red,221; green,132; blue,82}, opacity=0.2] (\xB,{\y - 0.1}) rectangle ({\xB + \linelen},{\y + 0.1});
            \node[anchor=west, font=\scriptsize] at ({\xB + \linelen + \textgap}, \y) {10 Agents};

            \draw[thick, color={rgb,255:red,85; green,168; blue,104}, opacity=0.8] (\xC,\y) -- ({\xC + \linelen},\y);
            \fill[color={rgb,255:red,85; green,168; blue,104}, opacity=0.2] (\xC,{\y - 0.1}) rectangle ({\xC + \linelen},{\y + 0.1});
            \node[anchor=west, font=\scriptsize] at ({\xC + \linelen + \textgap}, \y) {50 Agents};
        \end{tikzpicture}
    };
\end{tikzpicture}
    
  \begin{subfigure}{0.23\linewidth}
    \centering
    \includegraphics[width=\linewidth]{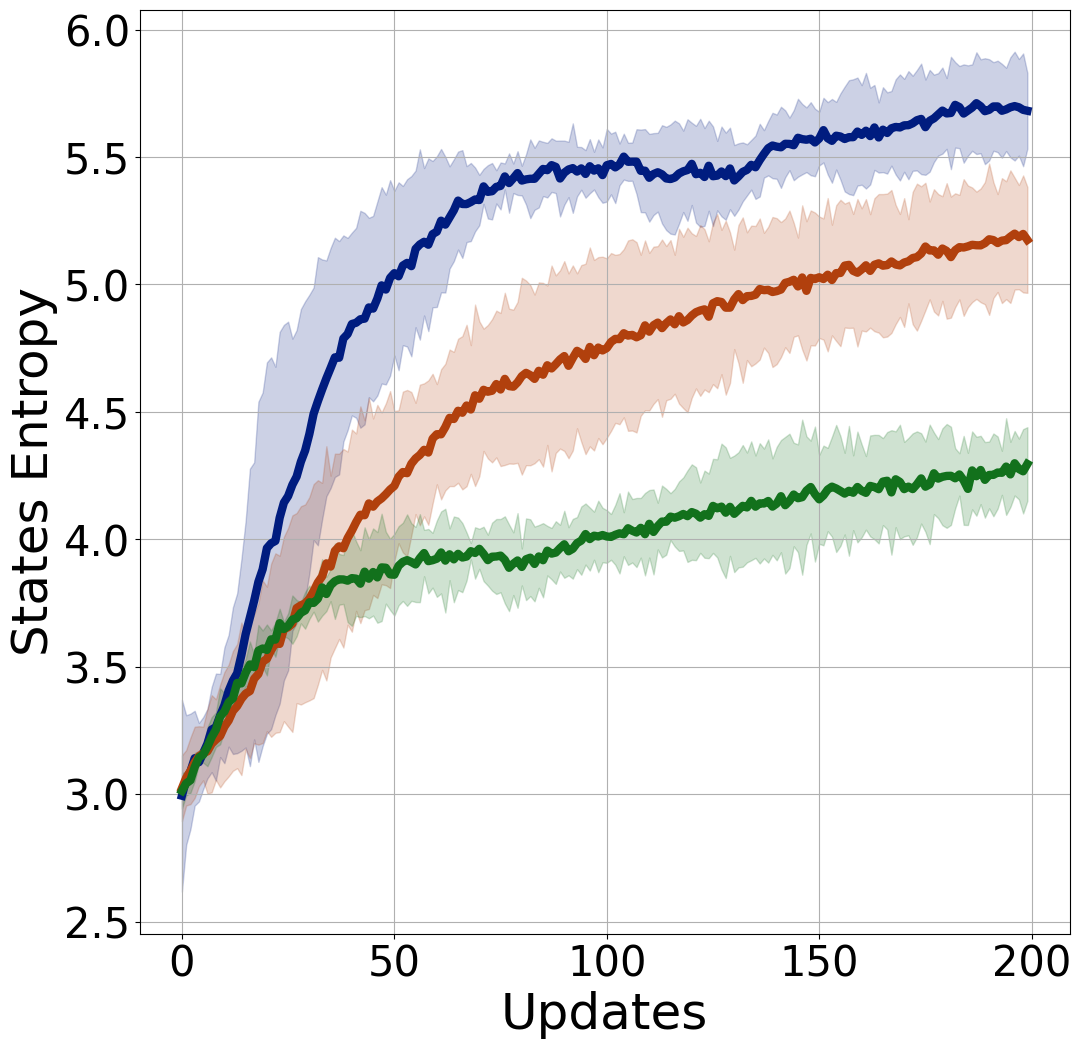}
    \caption{Ground Environment}
  \end{subfigure}
  \hfill
  \begin{subfigure}{0.23\linewidth}
    \centering
    \includegraphics[width=\linewidth]{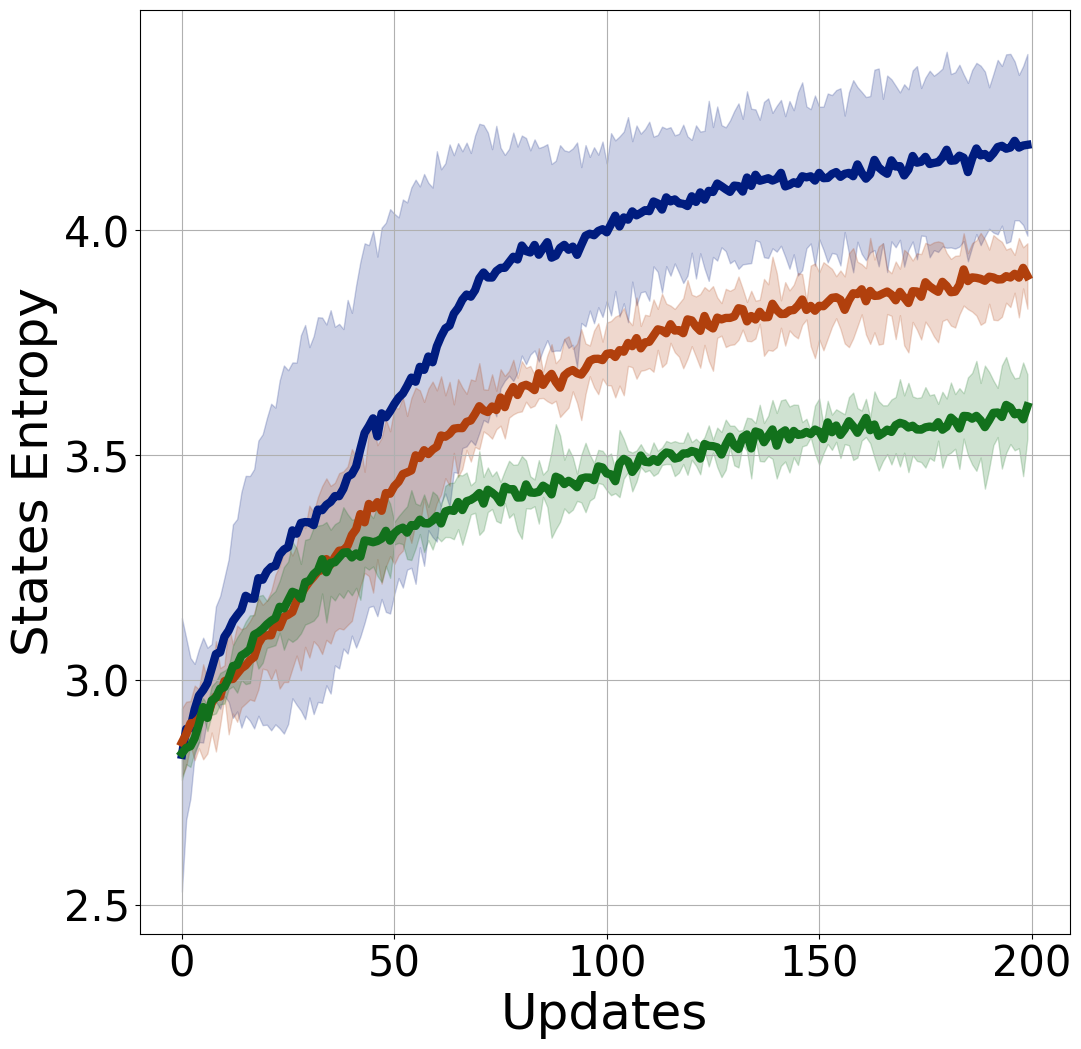}
    \caption{Maze Environment}
  \end{subfigure}
  \hfill
  \begin{subfigure}{0.23\linewidth}
    \centering
    \includegraphics[width=\linewidth]{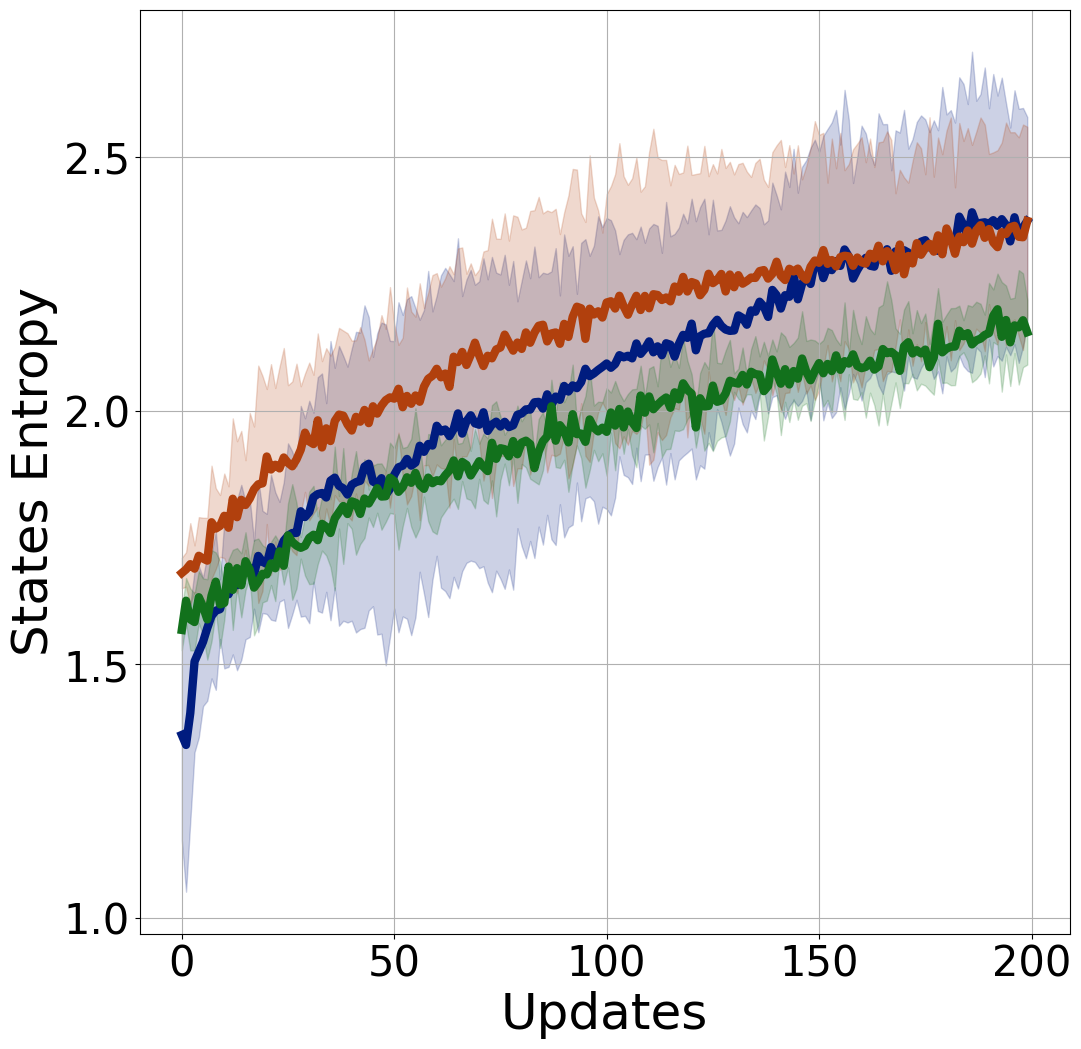}
    \caption{Cave Environment}
  \end{subfigure}
  \hfill
  \begin{subfigure}{0.23\linewidth}
    \centering
    \includegraphics[width=\linewidth]{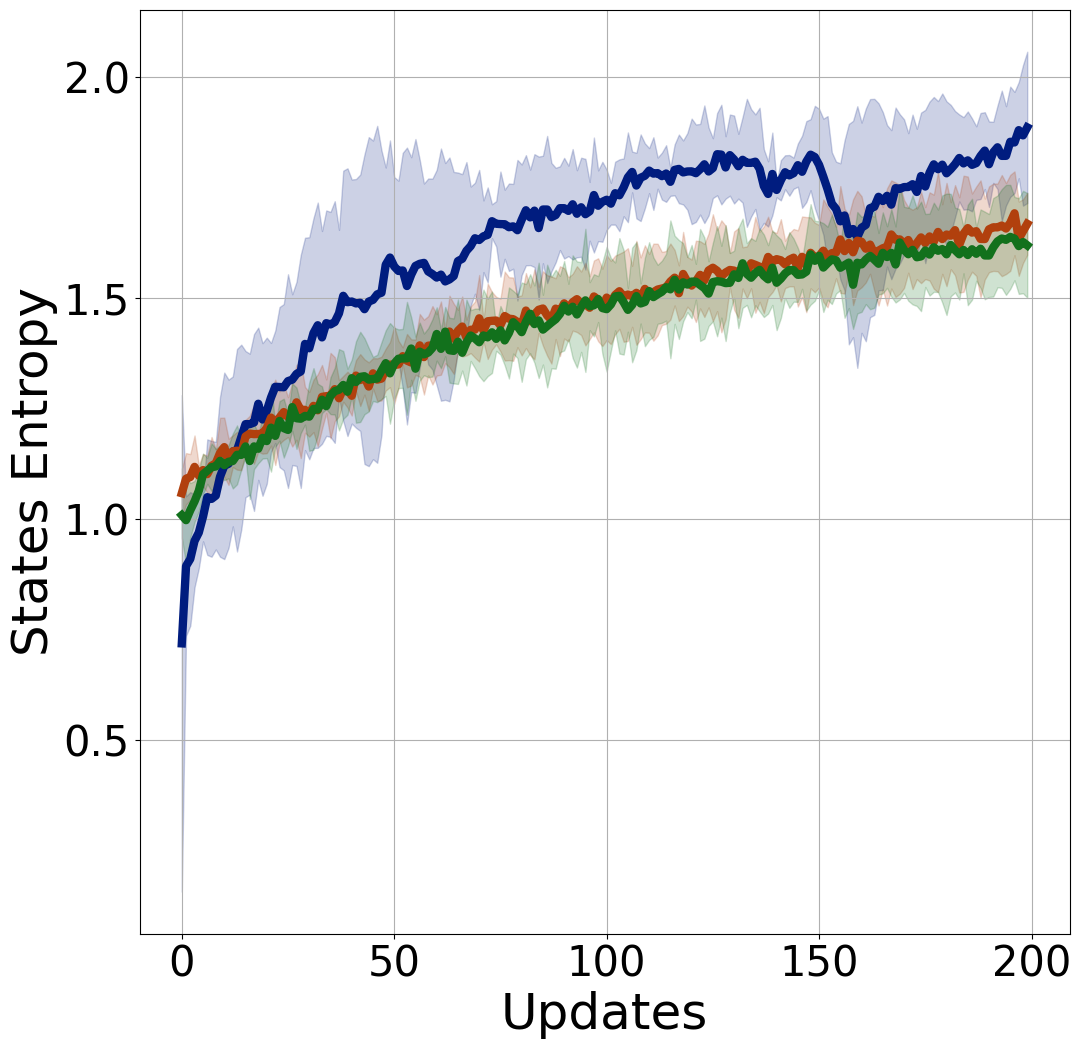}
    \caption{Pyramid Environment}
  \end{subfigure}

  \caption{Parallel state entropy achieved by a single generalist agent, against a collection of 10 or 50 agents. The plots report the average state entropy and 95\% c.i. across 4 runs as a function of the policy updates.}
  \label{fig:stateentropy-perfomance}
\end{figure*}

To highlight the advantages of designing algorithms that explicitly use parallelization at the algorithmic level, we present a set of experiments conducted with thousands of \emph{Ants} replicas in Isaac Lab \citep{mittal2023orbit}. The experiments are organized into two parts:
\begin{itemize}[topsep=0pt, itemsep=0pt, leftmargin=10pt]
    \item \textbf{Unsupervised Parallel Exploration}: We report the results of the learning process based on Algorithm~\ref{alg:kmyriad}, evaluating the performance of the learned policy as the number of agents scales.
    \item \textbf{Pretrained Locomotion}: We focus on the locomotion problem, analyzing the efficiency of pretraining parallel agents with heterogeneous policies derived from the optimization of~\ref{eq:mse_parallel}.
\end{itemize}
The list of environments used in the experiments is provided in detail in the supplementary materials. 
As a base environment, we instantiate a robotic ant whose final objective is to learn to navigate in the \emph{XY}-plane under various terrains and scenarios. The ant is a 3D quadruped robot consisting of a torso (a free-rotating body) with four legs attached, each leg composed of two segments. In the Ant implementation of Isaac Sim, the robot is controlled by applying torques to the joints connecting the legs to the torso. The observation space consists of $62$ features, capturing the position and velocity of the torso along with the contact forces experienced during locomotion.  
We design four environments of increasing difficulty: (i) an \emph{Empty} environment, where the ant can freely move in any direction; (ii) a \emph{Maze}, in which the ant must navigate corridors with collisions enabled to guide it along the path; and (iii–iv) two environments populated with \emph{Pyramids} and \emph{Caves}, respectively, which challenge the ant to traverse stairs and uneven terrain.

\subsection{Unsupervised Parallel Exploration}

To evaluate the performance of our proposed method, we conduct experiments within Isaac Sim using $1000$ \emph{Ant} instances. For the sake of reproducibility, experiments are repeated across multiple random seeds and with varying numbers of agents \( m \in \{1, 10, 50\} \). Each agent interacts with a subset of replicas through a policy network, implemented as a multi-head architecture, which is described in detail in Section \ref{sec:exploration}. 
From an experimental perspective, we set the number of trajectories such that $\tau \gg m$ to fully exploit Isaac Sim’s parallelization capabilities. This allows us to avoid looping over batches of trajectories and instead execute all rollouts in a single run, assigning each parallel agent to a subset of controlled replicas.

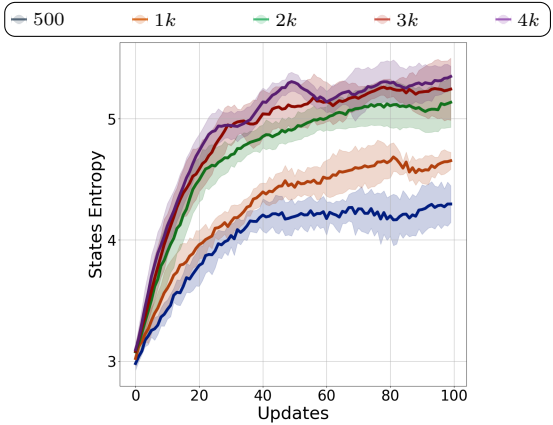
\begin{figure}[htbp]
    \centering
    \begin{tikzpicture}
    \node[draw=black, rounded corners, inner sep=2pt, fill=white] (legend) at (0,0) {
        \begin{tikzpicture}[scale=0.7]
            \def\linelen{0.3}
            \def\textgap{0.01}
            \def\y{0}

            \def\xA{0}
            \def\xB{2.3}
            \def\xC{4.6}
            \def\xD{6.9}
            \def\xE{9.2}

            \draw[thick, color={rgb,255:red,52; green,73; blue,94}, opacity=0.8] (\xA,\y) -- ({\xA + \linelen},\y);
            \fill[color={rgb,255:red,52; green,73; blue,94}, opacity=0.2] (\xA,{\y - 0.1}) rectangle ({\xA + \linelen},{\y + 0.1});
            \node[anchor=west, font=\scriptsize] at ({\xA + \linelen + \textgap}, \y) {$500$};

            \draw[thick, color={rgb,255:red,211; green,84; blue,0}, opacity=0.8] (\xB,\y) -- ({\xB + \linelen},\y);
            \fill[color={rgb,255:red,211; green,84; blue,0}, opacity=0.2] (\xB,{\y - 0.1}) rectangle ({\xB + \linelen},{\y + 0.1});
            \node[anchor=west, font=\scriptsize] at ({\xB + \linelen + \textgap}, \y) {$1k$};

            \draw[thick, color={rgb,255:red,39; green,174; blue,96}, opacity=0.8] (\xC,\y) -- ({\xC + \linelen},\y);
            \fill[color={rgb,255:red,39; green,174; blue,96}, opacity=0.2] (\xC,{\y - 0.1}) rectangle ({\xC + \linelen},{\y + 0.1});
            \node[anchor=west, font=\scriptsize] at ({\xC + \linelen + \textgap}, \y) {$2k$};

            \draw[thick, color={rgb,255:red,192; green,57; blue,43}, opacity=0.8] (\xD,\y) -- ({\xD + \linelen},\y);
            \fill[color={rgb,255:red,192; green,57; blue,43}, opacity=0.2] (\xD,{\y - 0.1}) rectangle ({\xD + \linelen},{\y + 0.1});
            \node[anchor=west, font=\scriptsize] at ({\xD + \linelen + \textgap}, \y) {$3k$};

            \draw[thick, color={rgb,255:red,142; green,68; blue,173}, opacity=0.8] (\xE,\y) -- ({\xE + \linelen},\y);
            \fill[color={rgb,255:red,142; green,68; blue,173}, opacity=0.2] (\xE,{\y - 0.1}) rectangle ({\xE + \linelen},{\y + 0.1});
            \node[anchor=west, font=\scriptsize] at ({\xE + \linelen + \textgap}, \y) {$4k$};

        \end{tikzpicture}
    };
\end{tikzpicture}
    \begin{subfigure}{\linewidth}
        \centering
        \includegraphics[width=0.62\linewidth]{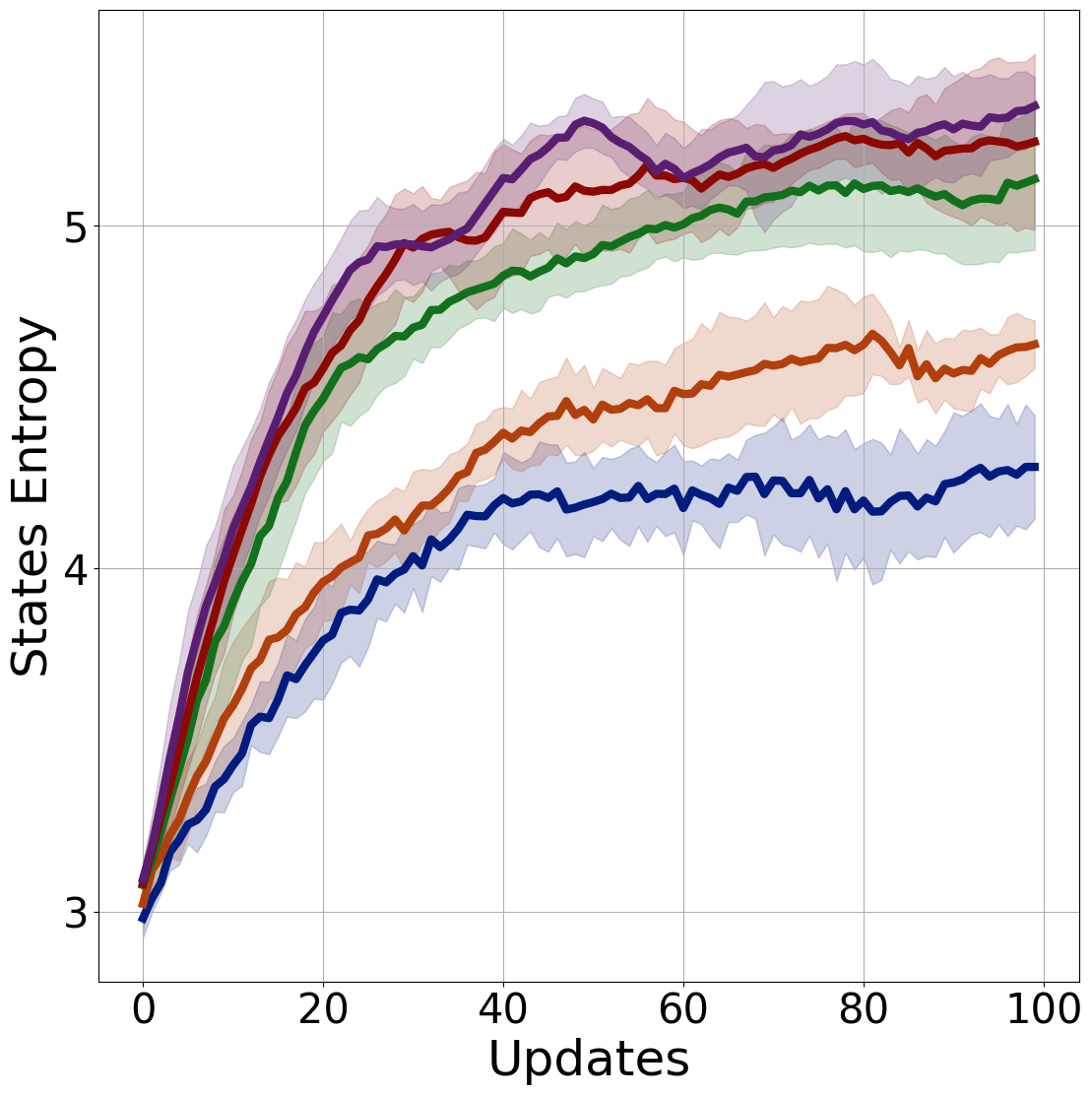}
    \end{subfigure}
    \caption{Increasing of the fixed overall trajectory ($\tau$) budget, over 4 runs, with 10 parallel learning agents.}
    \label{fig:peragentbudget}
\end{figure}

\begin{figure}[h]
    \centering
    \begin{subfigure}{\linewidth}
        \centering
        \includegraphics[width=\linewidth]{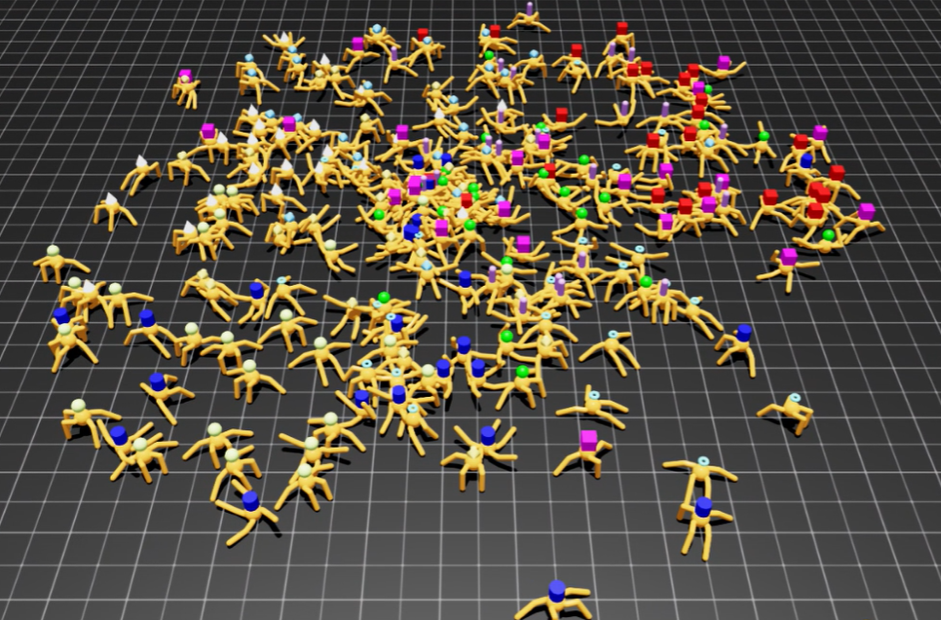}
    \end{subfigure}

    \caption{Diversity in state space visitation of \emph{Ant} replicas controlled by 10 independent policy heads.}
    \label{fig:diversity}
\end{figure}

Although the agents share parts of the policy network, they are treated as independent entities, each starting from the same initial state $s_0$ and evolving without interference during the simulation.
The training loop proceeds over a fixed number of epochs, during which trajectories with finite horizon are collected in parallel and used to update the policies based on Alg. \ref{alg:kmyriad}. Our reference baseline is a single agent controlling the entire set of replicas. We use this baseline to evaluate differences in the value of the achieved state entropy and the diversity of the obtained policies. In this setting, the single agent interacts with one thousand environments throughout the entire trajectory, thus benefiting from a huge batch size. To ensure fairness, we fix the total trajectory budget to match the number of replicas and progressively increase the number of agents. Consequently, each agent is updated using gradients computed from a smaller subset of trajectories, reflecting the trade-off between parallelism and per-agent sample efficiency.
In Figure~\ref{fig:stateentropy-perfomance}, we observe that for a fixed number of training episodes, the single agent achieves the highest entropy in environments where the ant can move freely. This apparent advantage stems from the larger batch size used for gradient updates. The group of 10 agents reaches entropy values close to the single-agent baseline, despite each agent having fewer samples with the population of 50 agents converges more slowly, since each policy is updated with gradients from an even smaller subset of trajectories. 
This slowdown is not fundamental but arises from dividing a fixed trajectory budget among multiple agents, reducing per-agent samples. Fig.~\ref{fig:peragentbudget} shows that increasing the overall budget (with $10$ agents, across $4$ seeds) improves state entropy and state-space coverage.In more complex environments such as \emph{pyramid} and \emph{cave}, however, this limitation is outweighed by the advantages of parallel exploration. Here, the 50-agent population discovers a broader repertoire of behaviors and achieves higher entropy with fewer samples. Considering playing the 10-parallel multi-head policy, as illustrated in Figure~\ref{fig:diversity}, this diversity emerges naturally from parallel optimization: individual agents tend to explore novel states while collectively maintaining balanced coverage. The resulting clusters, visible through distinct colors and markers, clearly highlight how each agent develops its own specialized policy.

Increasing the number of parallel agents enhances policy diversity. 
As empirical demonstration, Table~\ref{tab:diversity_kl} reports the KL divergence between parallel agents computed as follows: for each agent, we collect a batch of $1000$ trajectories and estimate the empirical KL divergence \cite{wang2009divergence} between the states it visits and those visited by all other agents. We then average the resulting divergences across agents.

\begin{figure*}[ht!] 
    \centering
    \begin{tikzpicture}
    \node[draw=black, rounded corners, inner sep=2pt, fill=white] (legend) at (0,0) {
        \begin{tikzpicture}[scale=0.7]
            \def\linelen{0.3}
            \def\linegap{0.05}
            \def\textgap{0.01}
            \def\y{0}

            \def\xA{0}
            \def\xB{2.5}
            \def\xC{5.0}

            \draw[thick, color=gray, opacity=0.8] (\xA,\y) -- ({\xA + \linelen},\y);
            \fill[color=gray, opacity=0.2] (\xA,{\y - 0.1}) rectangle ({\xA + \linelen},{\y + 0.1});
            \node[anchor=west, font=\scriptsize] at ({\xA + \linelen + \textgap}, \y) {Random};

            \draw[thick, color={rgb,255:red,76; green,114; blue,176}, opacity=0.8] (\xB,\y) -- ({\xB + \linelen},\y);
            \fill[color={rgb,255:red,76; green,114; blue,176}, opacity=0.2] (\xB,{\y - 0.1}) rectangle ({\xB + \linelen},{\y + 0.1});
            \node[anchor=west, font=\scriptsize] at ({\xB + \linelen + \textgap}, \y) {1 Agent};

            \draw[thick, color={rgb,255:red,85; green,168; blue,104}, opacity=0.8] (\xC,\y) -- ({\xC + \linelen},\y);
            \fill[color={rgb,255:red,85; green,168; blue,104}, opacity=0.2] (\xC,{\y - 0.1}) rectangle ({\xC + \linelen},{\y + 0.1});
            \node[anchor=west, font=\scriptsize] at ({\xC + \linelen + \textgap}, \y) {50 Agents};
        \end{tikzpicture}
    };
\end{tikzpicture}
    
  \begin{subfigure}{0.23\linewidth}
    \centering
    \includegraphics[width=\linewidth]{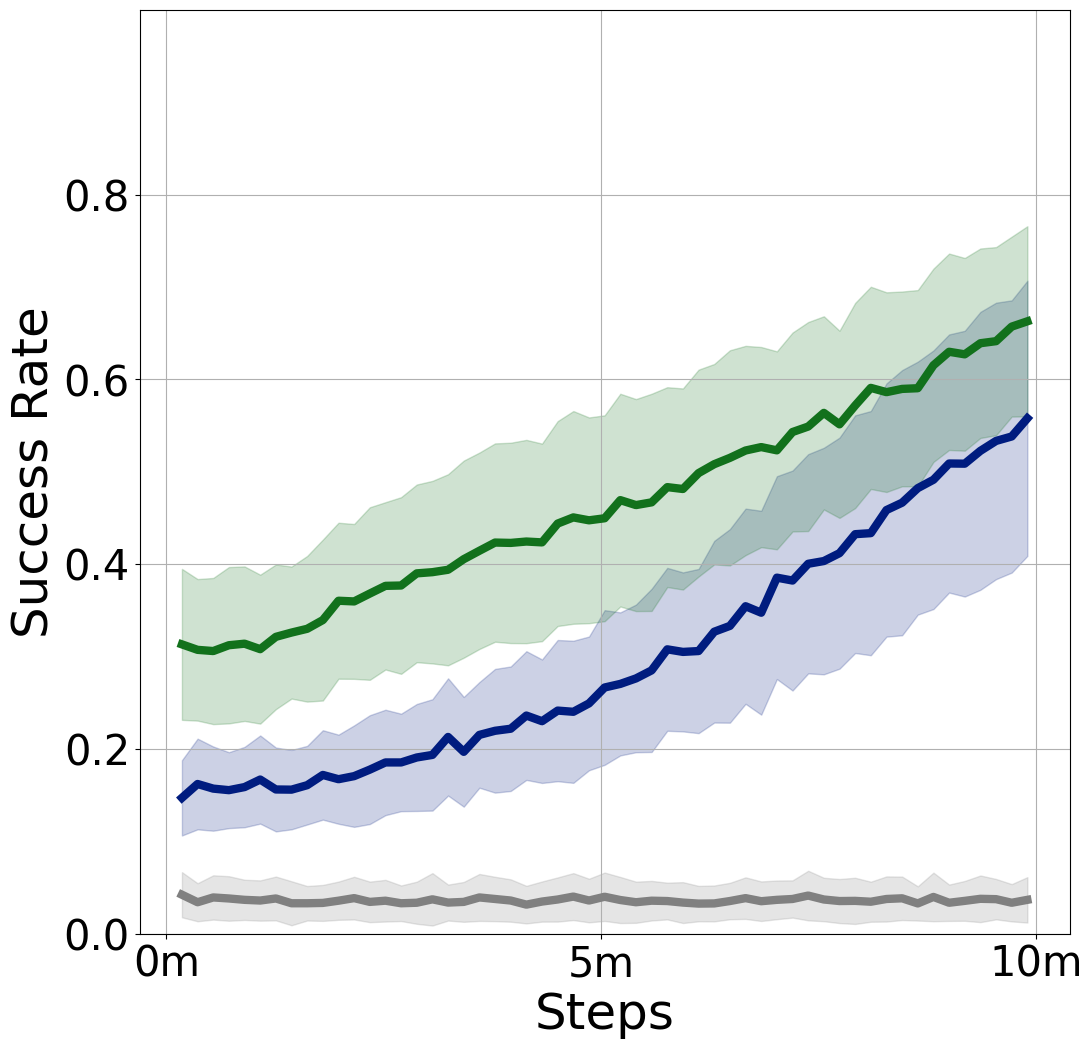}
    \caption{Ground Environment}
  \end{subfigure}
  \hfill
  \begin{subfigure}{0.23\linewidth}
    \centering
    \includegraphics[width=\linewidth]{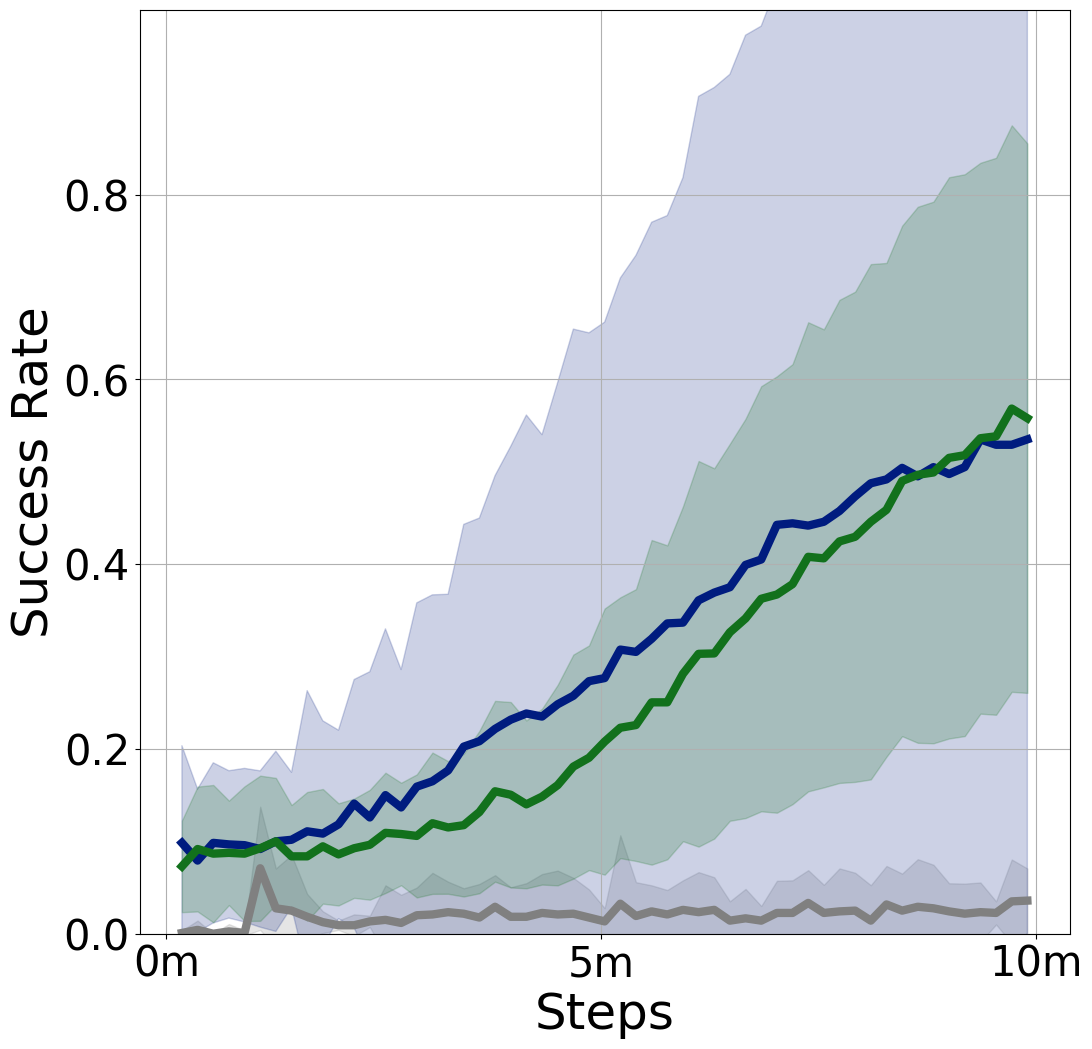}
    \caption{Maze Environment}
  \end{subfigure}
  \hfill
  \begin{subfigure}{0.23\linewidth}
    \centering
    \includegraphics[width=\linewidth]{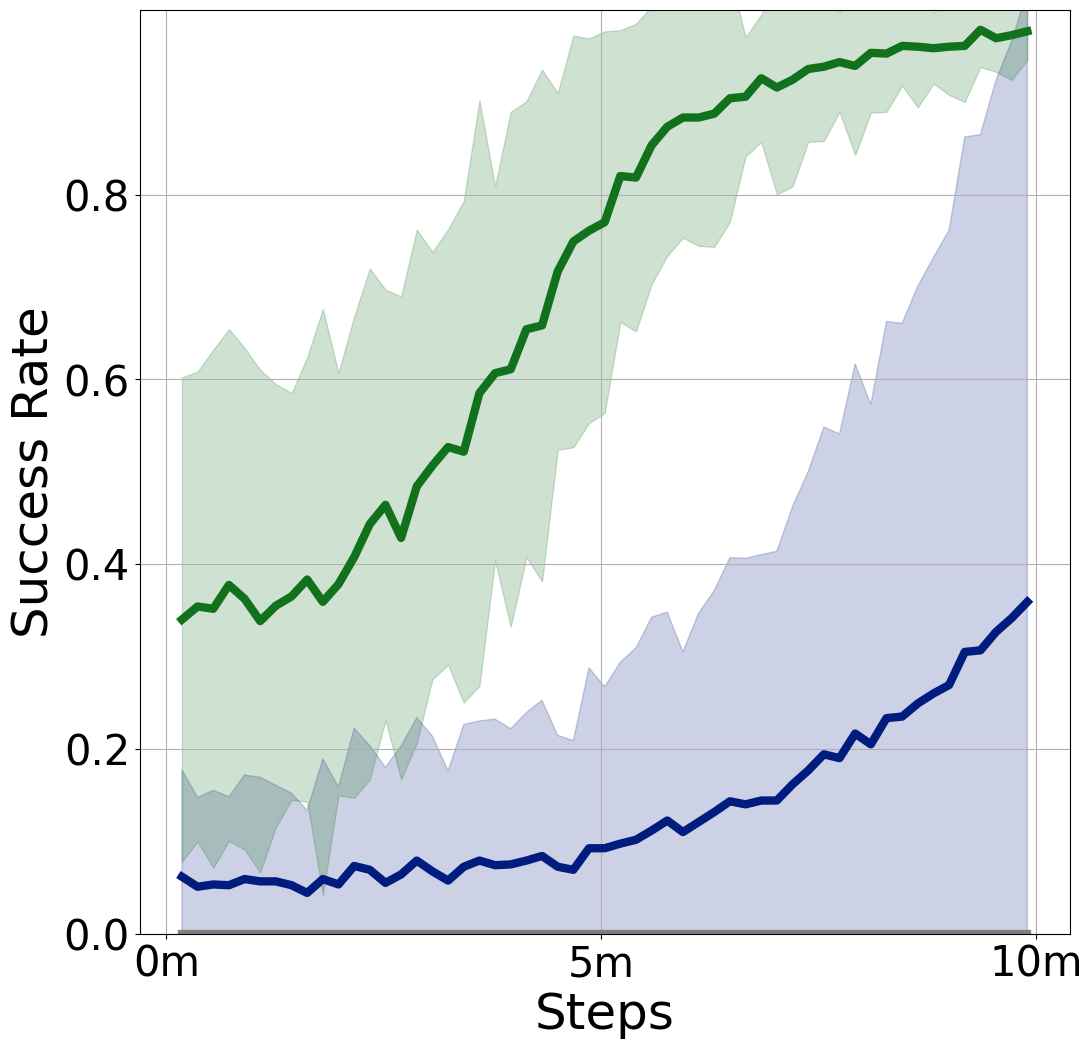}
    \caption{Cave Environment}
  \end{subfigure}
  \hfill
  \begin{subfigure}{0.23\linewidth}
    \centering
    \includegraphics[width=\linewidth]{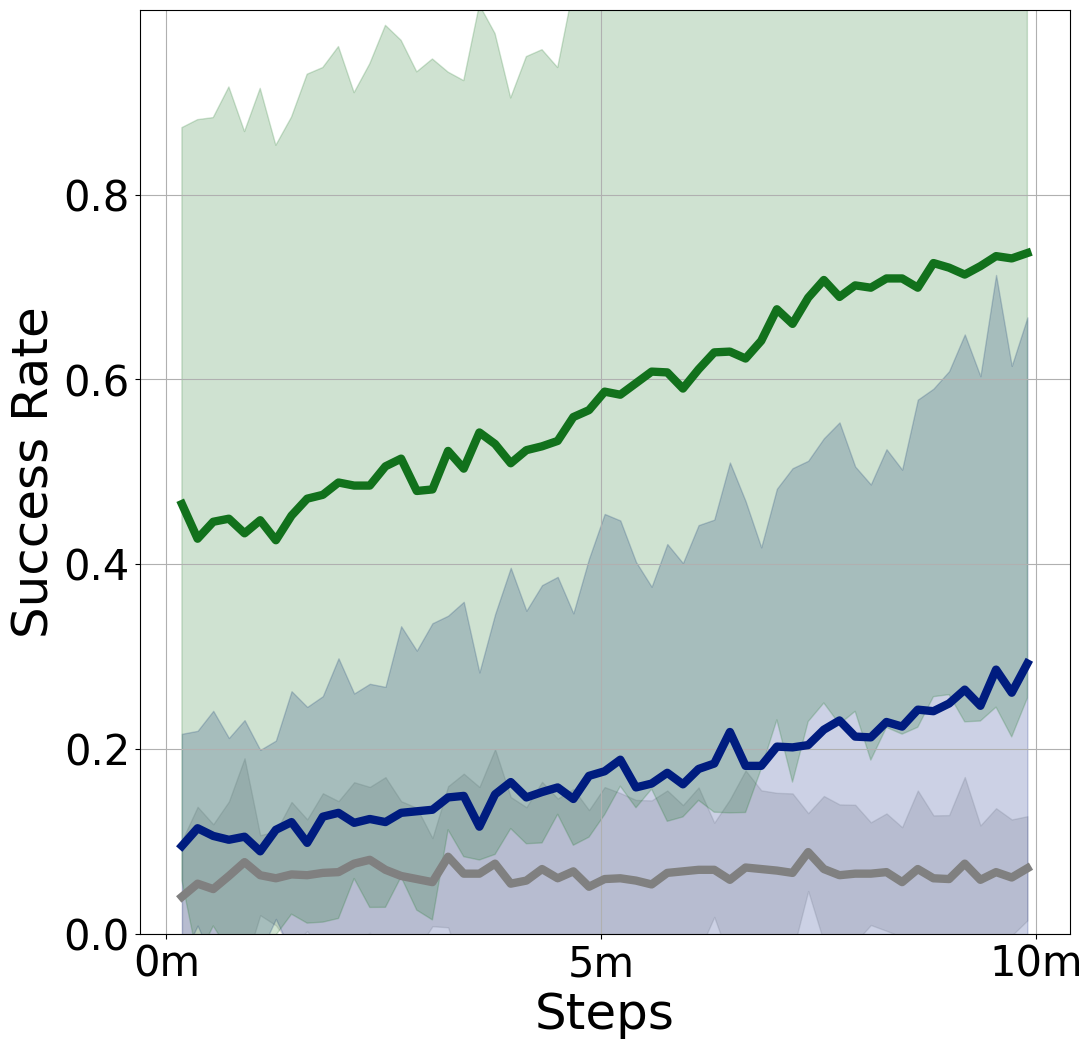}
    \caption{Pyramid Environment}
  \end{subfigure}

  \caption{PPO success rates, with different initialization strategies over multiple runs. Mean success rate over updates (solid line) with 95\% confidence interval across runs (shaded area)..}
  \label{fig:goalbasedperformance}
\end{figure*}
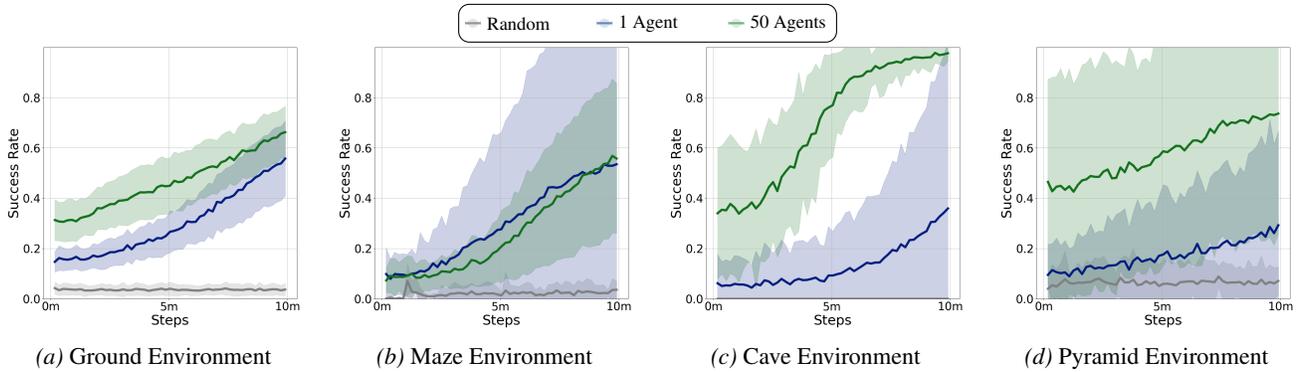

\begin{table}[ht]
\centering
\resizebox{\columnwidth}{!}{%
\begin{tabular}{lcccc}
\toprule
\textbf{Environment} & \textbf{Empty} & \textbf{Maze} & \textbf{Pyramid} & \textbf{Cave} \\
\midrule
\textbf{10 Agents} & $0.9869 \pm 0.0160$ & $0.5980 \pm 0.0267$ & $0.5340 \pm 0.0333$ & $0.8804 \pm 0.0399$ \\
\textbf{50 Agents} & \textbf{1.3337 $\pm$ 0.0099} & \textbf{1.0230 $\pm$ 0.0164} & \textbf{0.8807 $\pm$ 0.0288} & \textbf{1.0432 $\pm$ 0.0153} \\
\bottomrule
\end{tabular}
}
\caption{Empirical KL divergence between parallel agents.}
\label{tab:diversity_kl}
\end{table}

\subsection{Pretrained Locomotion}

To take advantage of the exploratory policies obtained with K-Myriad, we adopt a PPO \citep{schulman2017proximalpolicyoptimizationalgorithms} fine-tuning scheme with the following structure. 
In all the previous presented environments, we compare three baselines: (i) a PPO agent with an actor policy with random initialization, (ii) a PPO agent pretrained with a single-agent maximum-entropy policy, and (iii) a PPO agent initialized from the best among the 50 parallel agents pretrained as in the previous section. When pretrained policies are available, we evaluate multiple rollouts trajectories per head and select the head with the highest average success rate, converting it into a single-head actor. If no pretrained policies are available, we instead train multiple randomly initialized heads and retain the best-performing one. The reward function is sparse: agents receive a reward of $1$ if they enter a fixed radius around the goal, and $0$ otherwise. This setup reflects common practices in PPO usage while structuring a meaningful set of comparisons. PPO training proceeds with parallel rollouts of $300$ environment replicas, per update step. The results in Fig.~\ref{fig:goalbasedperformance} report success rates across goal locations, using the pre-trained initialization strategies described above. Goal locations are sampled within a fixed annular region around the starting state to ensure diverse task configurations ( see details in Appendix).  
Overall, parallel training yields better initialization, and the key factor is diversity. Indeed, the head used to pre-train the PPO agent already explores states close to the goal, which accelerates convergence by producing rollouts enriched with a high probability of goal-oriented trajectories. 
A minor drawback emerges near convergence, where the policy becomes more deterministic, leading to final performance comparable to that of single-agent pretraining. 
In contrast, the single-agent baseline starts with a highly entropic policy that slows down early learning but eventually benefits from richer rollouts, ultimately reaching an acceptable success rate. 
Crucially, within the limited interaction budget considered, both pretraining strategies clearly outperform a PPO agent with random initialization, which must first acquire basic locomotion before learning to reach the goal.

\section{Conclusions}
\label{sec:conclusions}

Our study shows that parallelism in reinforcement learning can serve not only to accelerate training, but also to enhance exploration by fostering behavioral diversity. By transforming parallel agents into a structured source of specialization, we demonstrate that pretrained policies can provide stronger initialization and faster adaptation than standard baselines.
We introduced \textbf{K-Myriad}, a scalable unsupervised pretraining framework that uses a shared-trunk, multi-head architecture to maximize collective state entropy. Unlike conventional parallelization methods that optimize a single policy, \textbf{K-Myriad} promotes differentiation across agents, yielding a portfolio of complementary exploration strategies.
Experiments in Isaac Sim show that this diversity improves performance on sparse-reward downstream tasks, while revealing a trade-off between policy diversity and per-agent sample efficiency under a fixed interaction budget.

\nocite{langley00}

\bibliography{example_paper}
\bibliographystyle{icml2026}

\newpage
\appendix
\onecolumn

\section{Environment Configurations}
\label{app:envs}

We design our experiments on top of \texttt{IsaacLab} \citep{mittal2023orbit}, using a manager-based 
environment with modular configuration classes. The environments differ 
primarily in the terrain setup, while sharing the same agent embodiment 
(the quadruped Ant robot) and task specification. 

\begin{figure}[ht]
    \centering
    \begin{subfigure}{0.45\linewidth}
        \centering
        \includegraphics[width=\linewidth]{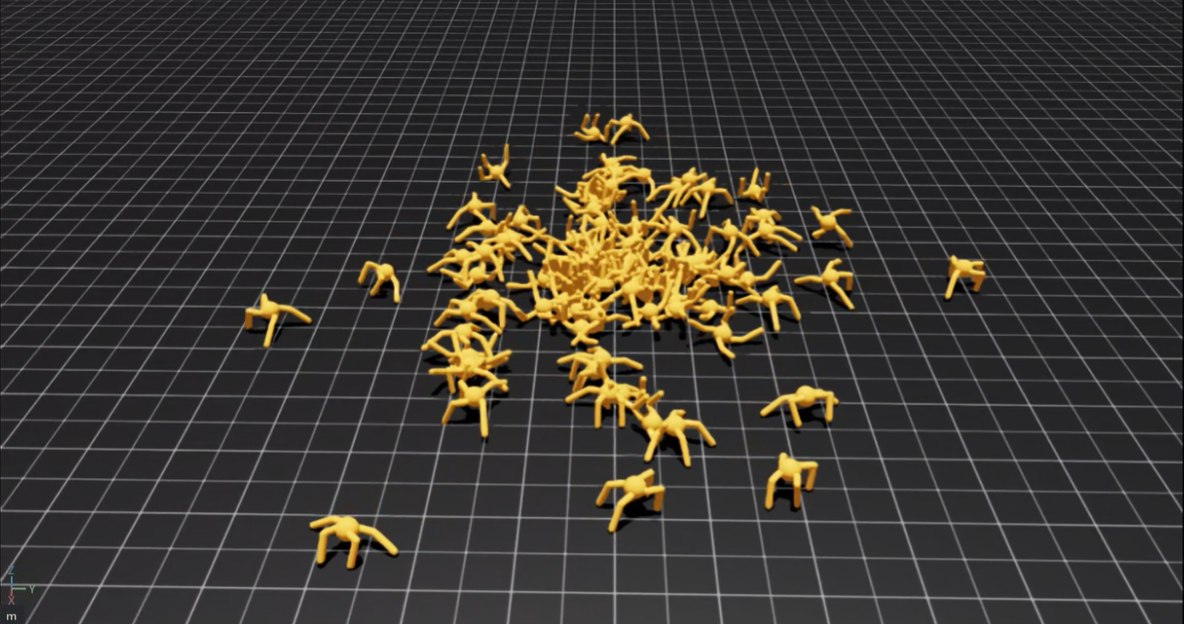}
        \caption{Scene with an Empty Environment}
        \label{fig:emptyenv}
    \end{subfigure}
    \hfill
    \begin{subfigure}{0.45\linewidth}
        \centering
        \includegraphics[width=\linewidth]{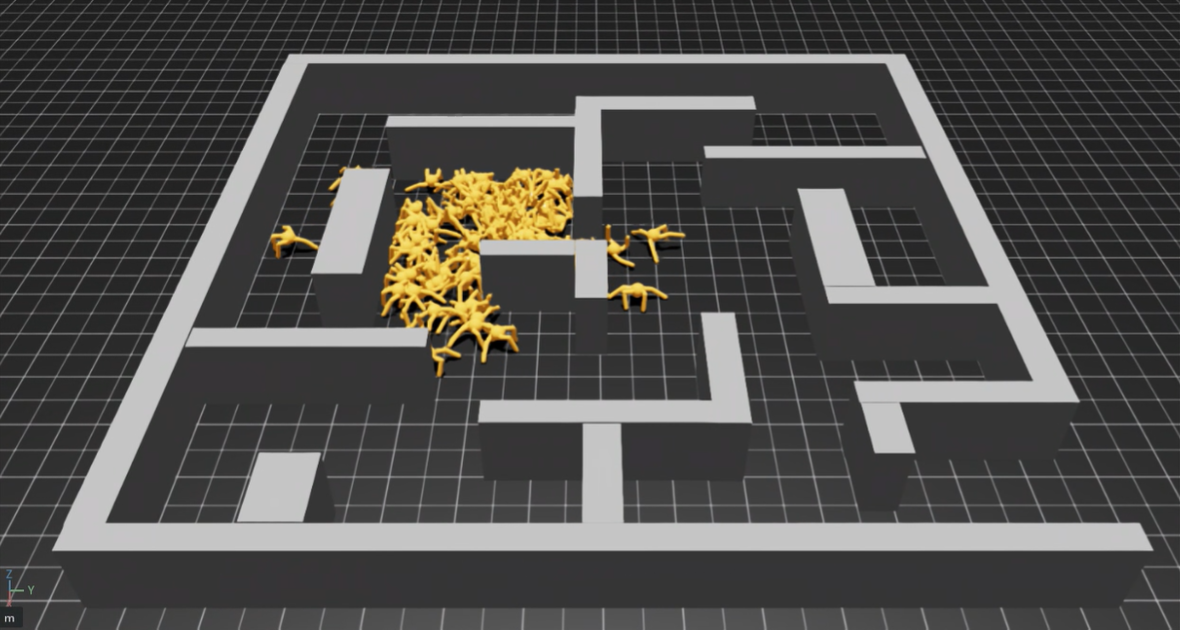}
        \caption{Scene with a Maze Environment}
        \label{fig:mazeenv}
    \end{subfigure}
    
    \vspace{0.5em} 

        \begin{subfigure}{0.45\linewidth}
        \centering
        \includegraphics[width=\linewidth]{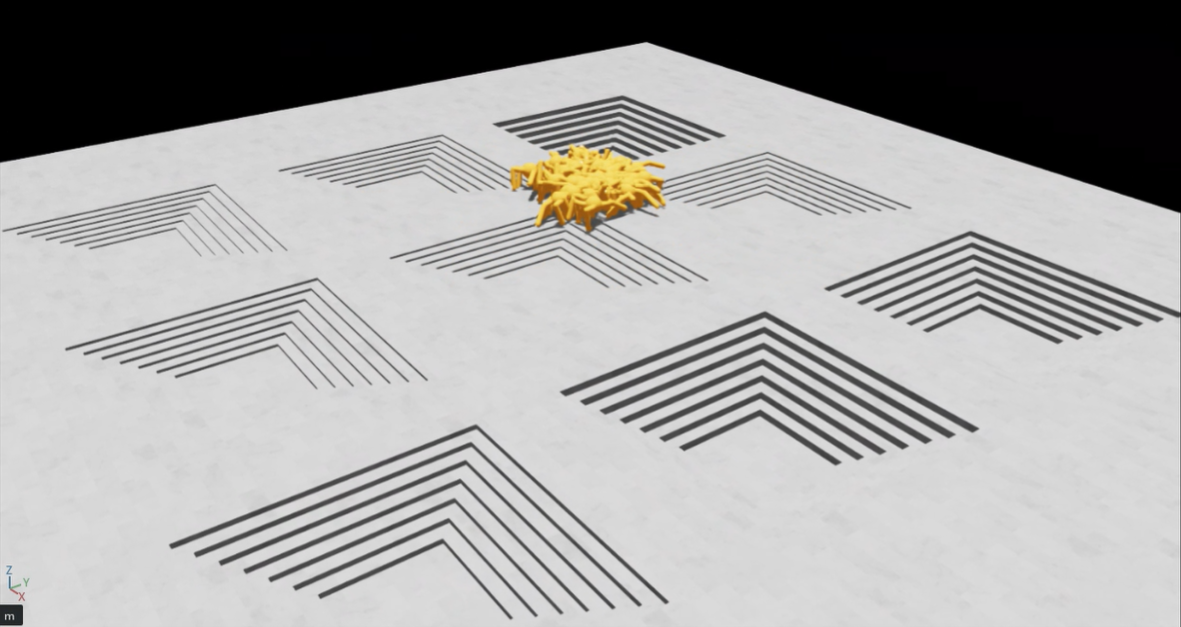}
        \caption{Scene full of Caves}
        \label{fig:caveenv}
    \end{subfigure}
    \hfill
    \begin{subfigure}{0.45\linewidth}
        \centering
        \includegraphics[width=\linewidth]{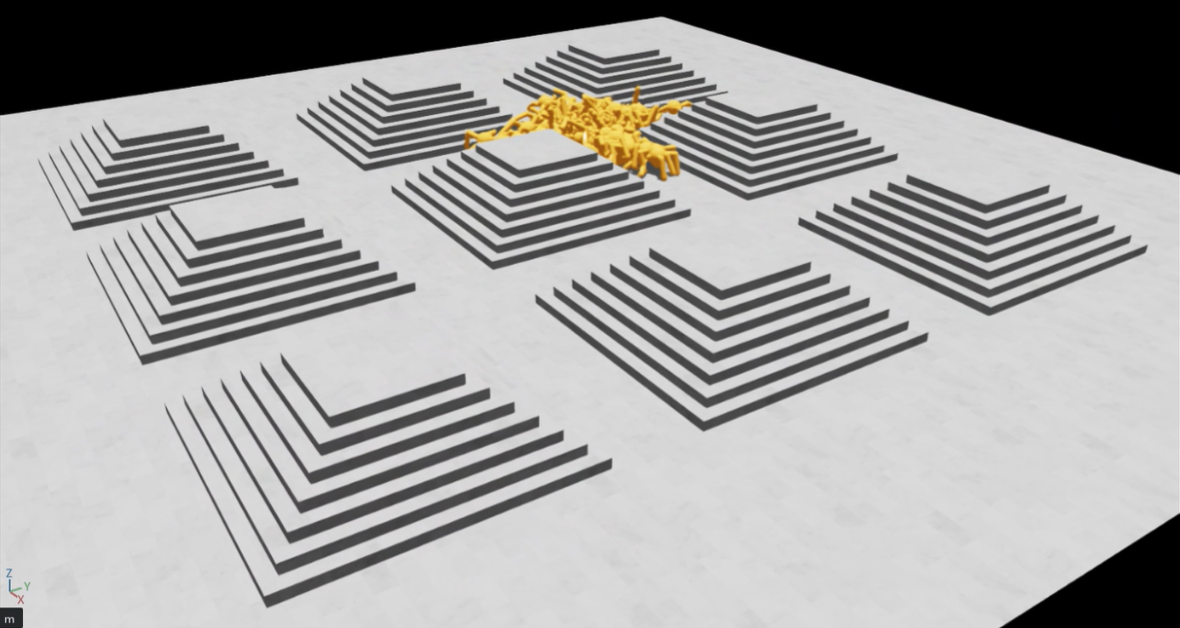}
        \caption{Scene full of Pyramids}
        \label{fig:pyramidenv}
    \end{subfigure}
    \caption{Overview of the environments used: (a) flat ground with an Ant robot, 
    (b) complex maze environment, (c) collection of pyramids, and 
    (d) cave-like terrain.}
    \label{fig:preliminaries}
\end{figure}

\subsection{Robot and Scene}
The agent is a simulated Ant robot placed in an 
interactive scene. Each environment spawns a terrain importer, with optional additional assets (e.g., a maze). 
The terrain determines the complexity of locomotion and exploration.

\paragraph{Empty Environment.} 
A flat plane is used as the terrain. The surface has standard physical 
properties, making it suitable for baseline locomotion experiments.

\paragraph{Maze Environment.} 
The terrain remains a flat plane, but a static maze structure 
is imported. The Ant must navigate corridors and 
obstacles, emphasizing exploration and spatial awareness.

\paragraph{Cave Environment.} 
The terrain is procedurally generated with irregular 
cave structures. This environment increases 
the difficulty of locomotion by requiring robust control to overcome 
slopes and lower surfaces.

\paragraph{Pyramid Environment.} 
The terrain is procedurally generated with irregular 
elevations and pyramid-like structures. This environment increases 
the difficulty of locomotion by requiring robust control to overcome 
slopes and uneven surfaces.

\subsection{Simulation Parameters}
Simulations run at a physics step of $\Delta t = 1/60$ seconds, 
with a control frequency given by a decimation factor of $5$. 
The physics engine uses PhysX settings with a bounce threshold 
velocity of $0.2$, static/dynamic friction set to $1.0$, and 
no restitution. All environments are highly parallelized, 
with up to $1000$ simulated instances in a single rollout.

\subsection{Action Space}
The agent controls all actuated joints of the Ant using an 
\textit{effort-based action space}. The action vector 
$a \in [-1, 1]^8$ is scaled by a factor of $7.5$ to generate 
joint torques. This low-level continuous action space ensures 
direct control over locomotion dynamics.

\subsection{Observations}
The policy receives a concatenation of state features, including:
\begin{itemize}
    \item Root position and height of the base.
    \item Linear and angular velocities.
    \item Yaw and roll orientation.
    \item Relative angle and heading projection with respect to a distant target.
    \item Normalized joint positions and relative joint velocities.
    \item Contact forces on all four feet.
    \item Previous actions.
\end{itemize}
These observations provide both proprioceptive and exteroceptive 
signals, allowing the policy to reason about stability, 
locomotion, and task progression. The shape of the observations thus results in [$1000,62$], the number of parallel replica environments with the corresponding ant observation size. Although the relative angle and heading projection with respect to a fixed distant target are not required for the maximum-entropy objective, we retain the default task configuration from the isaac sim repository to ensure reproducibility.

\subsection{Rewards}
The reward function is employed exclusively in the experiments evaluating the use of the exploratory policy for downstream tasks. Specifically, a sparse reward of $+1$ is assigned to the agent when it reaches the target position, defined as being within an Euclidean distance of less than $1\,\mathrm{m}$ from the goal in the environment described above.

\subsection{Termination Conditions}
In both experimental settings, the \textbf{reward-free} case 
(Section~\ref{sec:exploration}) and the \textbf{jump-starting} case (Section~\ref{sec:jumpstarting}), episodes terminate only when the maximum time horizon $T$ is reached.

\section{Policy Architecture}
\label{app:sharednetwork}

The policy architecture follows a \emph{shared trunk with agent-specific heads} design. 
A single feed-forward backbone network processes the input state into a shared 
latent embedding, which is then specialized for each agent via lightweight adapters 
and multi-head output layers. This design enables parameter sharing across agents 
while still allowing for agent-specific action distributions.

\paragraph{Shared Trunk.} 
The shared trunk is a multilayer perceptron (MLP) with ReLU activations:
\[
x = f_\theta(s), \quad f_\theta: \mathbb{R}^{d_s} \to \mathbb{R}^{d_h},
\]
where $s$ is the input state, $d_s$ the state dimension, and $d_h$ the hidden size 
of the last trunk layer. This embedding $x$ is common to all agents.

\paragraph{Agent-Specific Adapters.}
For each agent $i \in \{1, \dots, m\}$, an adapter network 
\[
z_i = g_i(x), \quad g_i: \mathbb{R}^{d_h} \to \mathbb{R}^{d_a}
\]
projects the shared embedding into an agent-specific representation. 
This lightweight specialization layer allows agents to adapt the shared 
features to their own behavioral characteristics.

\paragraph{Multi-Head Outputs.}
The final action distribution is produced by vectorized multi-head linear 
layers. For each agent $i$, we compute:
\[
\mu_i = W^{(\mu)}_i z_i + b^{(\mu)}_i, \qquad 
\log\sigma_i = W^{(\sigma)}_i z_i + b^{(\sigma)}_i,
\]
where $W^{(\mu)}_i, W^{(\sigma)}_i$ and biases are head-specific parameters. 
The outputs $(\mu_i, \sigma_i)$ parameterize a Gaussian policy 
$\pi_i(a|s) = \mathcal{N}(\mu_i, \sigma_i^2)$, followed by a 
$\tanh$-squashing and affine rescaling to match the action space.

\paragraph{Importance of Sharing.}
By sharing the trunk $f_\theta$ across all agents, the network exploits 
common structure in the dynamics and observations, while the per-agent heads 
preserve diversity.

\paragraph{Loss Calculation}\label{sec:LossCalculation}

We now describe how we compute gradients of the maximum-entropy objective in Eq.~\ref{eq:maxentropyestimator} with respect to the parameters of the parallel policies. Directly differentiating entropy over high-dimensional state distributions is intractable; therefore, following prior work, we adopt a non-parametric $k$-NN entropy estimator augmented with \emph{importance weighting} \cite{AJGL201111991}. This formulation enables principled gradient computation with respect to policy parameters while remaining sample efficient.

The resulting entropy estimator is given by:
\begin{equation}
\label{eq:unsupervisedobjective}
\hat H_k(d_{\pi_p}) = - \frac{1}{N} \sum_{j=1}^N
\Bigg[ W_j \log \frac{W_j}{V_j} \Bigg] + \log k - \Psi(k),
\end{equation}
where $W_j$ denotes the aggregated importance weight of the $k$ nearest neighbors of sample $s_j$.

To maximize this entropy estimate, we apply an importance-weighted policy gradient that corrects for the mismatch between the behavioral policy $\beta_i$, under which data are collected, and the target policy $\pi_i$, which is being optimized. In our on-policy setting, $\beta_i$ is defined as a snapshot of $\pi_i$ at the beginning of each rollout phase, ensuring that updates remain theoretically unbiased despite policy evolution during optimization.

The per-trajectory importance weight is computed as:
\begin{equation}
\label{eq:IW}
w_t = \exp\left( \sum_{u=0}^t \log \pi_i(a_u \mid s_u) - \log \beta_i(a_u \mid s_u) \right).
\end{equation}

In our implementation, the behavioral policy coincides with the policy that generated each sampled particle. As a result, all importance weights evaluate to one, making the estimator numerically equivalent to the standard (unweighted) $k$-NN entropy estimator. This preserves a principled gradient formulation while simplifying computation in practice.

\section{Additional Experiments Details}

\paragraph{Reward free hyperparameters}

Table~\ref{tab:rf_hyperparams} summarizes the task and algorithm configurations used for training in the single-, 10-, and 50-agent experiments described in Section~\ref{sec:experiments}.
The single-agent baseline shares the same random seed, total number of trajectories, and trajectory length as the multi-agent setups, but differs in the number of trajectories processed per update and in the structure of the policy head.

\paragraph{State Visitation Heatmaps}
Figure~\ref{fig:heatmaps} reports the collection of state-visitation heatmaps obtained at the final training iteration for the different pretraining strategies described in Section~\ref{sec:experiments}. These visualizations correspond closely to the entropy trends shown in Figure~\ref{fig:stateentropy-perfomance}, providing spatial insight into the most frequently visited regions across environments.
In the simpler \emph{Empty} and \emph{Maze} environments, the single-agent and 10-agent versions of K-Myriad exhibit broader exploration and higher performance compared to the 50-agent implementation. However, this ability to explore diverse states is significantly reduced in more complex settings such as \emph{Pyramid} and \emph{Cave}. In these environments, successful behavior requires not only expanding coverage in the X–Y plane but also along the Z-axis, such as climbing the pyramid or descending into the cave, resulting in higher-performing agents that achieve a wider spatial perimeter of visited states.

\paragraph{Goal based experiments hyperparameters}
Table~\ref{tab:ppo_hyperparams} lists the PPO hyperparameters and corresponding configurations used in Section~\ref{sec:experiments} for comparing the performance of the different pretraining strategies. The parameters remain mostly consistent across settings, except for the policy learning rate, entropy coefficient, and clipping range, which are reduced in the pretrained cases to mitigate catastrophic forgetting in the reinforcement learning agents.

\renewcommand{\arraystretch}{1.3} 
\begin{table}[t]
    \centering
    \caption{\texttt{Reward-Free} K-myriad parameters}
    \vspace{-0.2cm}
    \resizebox{\textwidth}{!}{
    \begin{tabular}{l c r r r}
        \hline
        \textbf{Parameter} & \textbf{Symbol} & \texttt{Single Agent} & \texttt{10 Agents} & \texttt{50 Agents} \\
        \hline
        Number of agents & $a$ & $1$ & $10$ & $50$ \\
        Total trajectories (overall) & $N_{\mathrm{traj}}$ & $1000$ & $1000$ & $1000$ \\
        Number of environments replica & $N_{\mathrm{env}}$ & $1000$ & $1000$ & $1000$ \\
        Trajectories per agent & $N_{\mathrm{traj}}/a$ & $1000$ & $100$ & $20$ \\
        Trajectory length & $T$ & $600$ & $600$ & $600$ \\
        Seeds & $\mathcal{S}$ & $\{0,1,56,123\}$ & $\{0,1,56,123\}$ & $\{0,1,56,123\}$ \\
        Policy architecture (Trunk hidden sizes) & -- & $[512,\,256]$ & $[512,\,256]$ & $[512,\,256]$ \\
        Policy architecture (Head adapters) & -- & $1 \times[\,256,\,8]$ & $10 \times[\,256,\,8]$ & $50 \times[\,256,\,8]$ \\
        Learning rate & $\alpha$ & $2\times10^{-4}$ & $2\times10^{-4}$ & $5\times10^{-4}$ \\
        LR scheduler (milestones, $\gamma$) & -- & $[30,\,80],\ 0.5$ & $[30,\,80],\ 0.5$ & $[30,\,80],\ 0.5$ \\
        k-NN (KNN step) & $k$ & $5$ & $5$ & $5$ \\
        \hline
    \end{tabular}
    }
    \label{tab:rf_hyperparams}
\end{table}

\renewcommand{\arraystretch}{1.3} 
\begin{table}[t]
    \centering
    \caption{\texttt{PPO} parameters}
    \vspace{-0.2cm}
    \resizebox{\textwidth}{!}{
    \begin{tabular}{l c r r r}
        \hline
        \textbf{Parameter} & \textbf{Symbol} & \texttt{No Pretrain} & \texttt{1-Agent Pretrain} & \texttt{50-Agents Pretrain} \\
        \hline
        Policy hidden sizes & $H_{\pi}$ & $[512,\,256,\,256]$ & $[512,\,256,\,256]$ & $[512,\,256,\,256]$ \\
        Value function hidden sizes & $H_{V}$ & $[256,\,256]$ & $[256,\,256]$ & $[256,\,256]$ \\
        Policy learning rate & $\mathrm{LR}_{\pi}$ & $1\times10^{-5}$ & $1\times10^{-5}$ & $1\times10^{-5}$ \\
        Value function learning rate & $\mathrm{LR}_{V}$ & $3\times10^{-4}$ & $3\times10^{-4}$ & $3\times10^{-4}$ \\
        Discount factor & $\gamma$ & $0.99$ & $0.99$ & $0.99$ \\
        GAE parameter & $\lambda$ & $0.95$ & $0.95$ & $0.95$ \\
        Clipping range & $\epsilon$ & $0.20$ & $0.15$ & $0.15$ \\
        Entropy coefficient & $\beta_{\text{ent}}$ & $0.001$ & $0.0$ & $0.0$ \\
        Value loss coefficient & $\beta_{V}$ & $0.5$ & $0.5$ & $0.5$ \\
        Max gradient norm & $\|\nabla\|_{\max}$ & $0.5$ & $0.5$ & $0.5$ \\
        PPO epochs per rollout & $N_{\text{epoch}}$ & $10$ & $3$ & $3$ \\
        Warmup Rollout  &  & $1$ & $5$ & $5$ \\
        Minibatch size & $B_{\text{min}}$ & $64* num\_envs$ & $64* num\_envs$ & $64* num\_envs$ \\
        Rollout length & $T$ & $600$ & $600$ & $600$ \\
        Number of environments replica & $N_{\mathrm{env}}$ & $300$ & $300$ & $300$ \\
        Total timesteps & $N_{\text{steps}}$ & $1\times10^{7}$ & $1\times10^{7}$ & $1\times10^{7}$ \\
        \hline
    \end{tabular}
    }
    \label{tab:ppo_hyperparams}
\end{table}

\paragraph{Computational Resources}
All experiments were conducted on a  local workstation equipped with a 12th Gen Intel(R) Core(TM) i9-12900E CPU, 64\,GiB of RAM, and an NVIDIA RTX~5000 GPU (24\.GiB). To speed up experiments, we in addition work on cloud workstations provisioned through Brev,  equipped with a single NVIDIA A6000 GPU (48\,GiB), 28 CPU cores, and 58\,GiB of system memory, running on Hyperstack infrastructure. Together, these environments provided sufficient GPU memory and parallel compute capabilities to train and evaluate reinforcement learning agents at scale. With the introduced computational architecture we report the wall clock time for the experiments provided:
\begin{itemize}
    \item Maximum Entropy Empty Environment: $200\ epochs,1000\ replicas,120M\ Timesteps,[1,10,50]\ agents$ approximately \emph{40min} per run
    \item Maximum Entropy Maze Environment: $200\ epochs,1000\ replicas,120M\ Timesteps,[1,10,50]\ agents$ approximately \emph{51min} per run
    \item Maximum Entropy Pyramid Environment: $200\ epochs,1000\ replicas,120M\ Timesteps,[1,10,50]\ agents$ approximately \emph{48min} per run
    \item Maximum Entropy Cave Environment: $200\ epochs,1000\ replicas,120M\ Timesteps,[1,10,50]\ agents$ approximately \emph{44min} per run
    \item Maximum Entropy Empty Environment: $1000\ epochs,1000\ replicas,600M\ Timesteps,[1,10,50]\ agents$ approximately \emph{3h} per run.
    \item Goal Based PPO training Empty Environment:  $300\ replicas,10M\ Timesteps,[no-pretrain,1,50]\ agents$ approximately \emph{25min} per run.
    \item Goal Based PPO training Empty Environment:  $300\ replicas,15M\ Timesteps,[no-pretrain]\ agents$ approximately \emph{40min} per run.
\end{itemize}

\begin{figure}[h]
\centering
 \begin{subfigure}{0.4\linewidth}
    \centering
    \begin{tikzpicture}
    \node[draw=black, rounded corners, inner sep=2pt, fill=white] (legend) at (0,0) {
        \begin{tikzpicture}[scale=0.7]
            \def\linelen{0.3}
            \def\linegap{0.05}
            \def\textgap{0.01}
            \def\y{0}

            \def\xA{0}
            \def\xB{2.5}
            \def\xC{5.0}

            \draw[thick, color={rgb,255:red,76; green,114; blue,176}, opacity=0.8] (\xA,\y) -- ({\xA + \linelen},\y);
            \fill[color={rgb,255:red,76; green,114; blue,176}, opacity=0.2] (\xA,{\y - 0.1}) rectangle ({\xA + \linelen},{\y + 0.1});
            \node[anchor=west, font=\scriptsize] at ({\xA + \linelen + \textgap}, \y) {1 Agents};

            \draw[thick, color={rgb,255:red,221; green,132; blue,82}, opacity=0.8] (\xB,\y) -- ({\xB + \linelen},\y);
            \fill[color={rgb,255:red,221; green,132; blue,82}, opacity=0.2] (\xB,{\y - 0.1}) rectangle ({\xB + \linelen},{\y + 0.1});
            \node[anchor=west, font=\scriptsize] at ({\xB + \linelen + \textgap}, \y) {10 Agents};

            \draw[thick, color={rgb,255:red,85; green,168; blue,104}, opacity=0.8] (\xC,\y) -- ({\xC + \linelen},\y);
            \fill[color={rgb,255:red,85; green,168; blue,104}, opacity=0.2] (\xC,{\y - 0.1}) rectangle ({\xC + \linelen},{\y + 0.1});
            \node[anchor=west, font=\scriptsize] at ({\xC + \linelen + \textgap}, \y) {50 Agents};
        \end{tikzpicture}
    };
\end{tikzpicture}
    \centering
        \includegraphics[width=\linewidth]{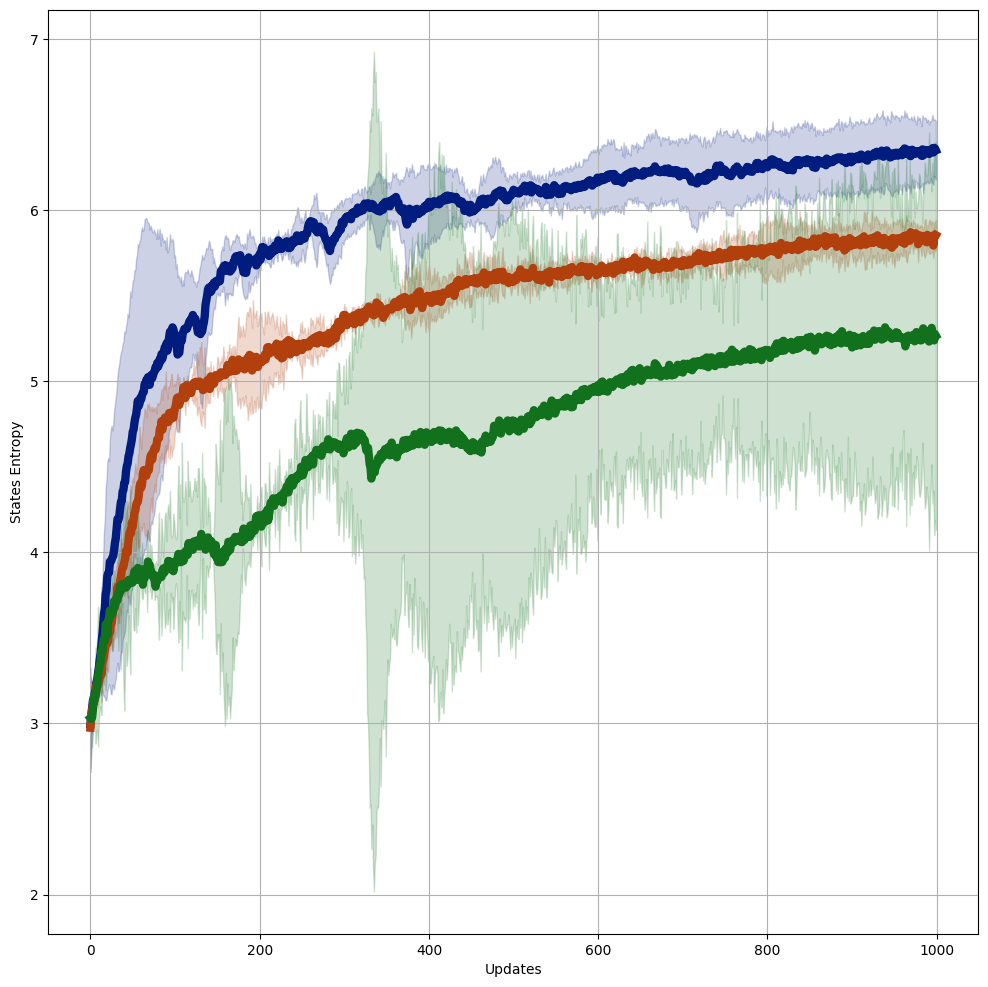}
        
    \end{subfigure}
    \caption{Parallel state entropy achieved by a single maximum entropy agent, against a collection of 10 or 50 agents, over longer runs. Mean State Entropy (solid line) and 95 \% c.i. (shaded area)}
    \label{fig:ablation}
\end{figure}

\subsection{Ablation on Pretraining and Agent Scaling}

\paragraph{Convergence of Parallel agent entropy performance}
From Figure~\ref{fig:ablation}, we analyze the entropy evolution over an extended training run of 1,000 update steps in the \emph{Empty} environment, which exhibits the most pronounced differences among the evaluated configurations. The single-agent setup continues to outperform the parallel-agent configurations, although the performance gap progressively narrows with training. This trend supports the hypothesis that, given sufficient updates, the parallel configurations tend to converge toward the behavior of the single-agent case.

\paragraph{PPO Behavior Across Goal Regions}
The results in Figure~\ref{fig:goal-success-comparison} highlight a clear difference in the PPO agent's performance depending on the goal location. The agent demonstrates consistent improvement and achieves high success rates for goals located within the affordable region (i.e., $|goal| < r$) with radius of $4.5 m$. In contrast, the performance remains poor for the original goal set introduced in Section~\ref{sec:experiments}, corresponding to targets farther from the initial state ($|goal| > r$), even after extended training. This outcome indicates that PPO alone struggles to generalize to distant goals, and pretraining is therefore essential to effectively solve tasks that lie outside the agent's initial reachable area.

\begin{figure}
    \begin{subfigure}{0.4\linewidth}
        \centering
        \begin{tikzpicture}
    \node[draw=black, rounded corners, inner sep=2pt, fill=white] (legend) at (0,0) {
        \begin{tikzpicture}[scale=0.7]
            \def\linelen{0.3}
            \def\textgap{0.1}
            \def\yA{0}
            \def\yB{-0.5}
            \def\yC{-1.0}

            \filldraw[color={rgb,255:red,76; green,114; blue,176}] (0,\yA) circle (2pt);
            \node[anchor=west, font=\scriptsize] at ({\linelen + \textgap}, \yA) {Section \ref{sec:experiments} Goals};

            \draw[line width=0.7pt, color={rgb,255:red,255; green,191; blue,0}] 
                (-0.07,\yB-0.07) -- (0.07,\yB+0.07);
            \draw[line width=0.7pt, color={rgb,255:red,255; green,191; blue,0}] 
                (-0.07,\yB+0.07) -- (0.07,\yB-0.07);
            \node[anchor=west, font=\scriptsize] at ({\linelen + \textgap}, \yB) {New Goals};

            \draw[green, dashed, thick] (0,\yC) circle (3pt);
            \node[anchor=west, font=\scriptsize] at ({\linelen + \textgap}, \yC) {affordable area};
        \end{tikzpicture}
    };
\end{tikzpicture}
        \centering
        \includegraphics[width=\linewidth]{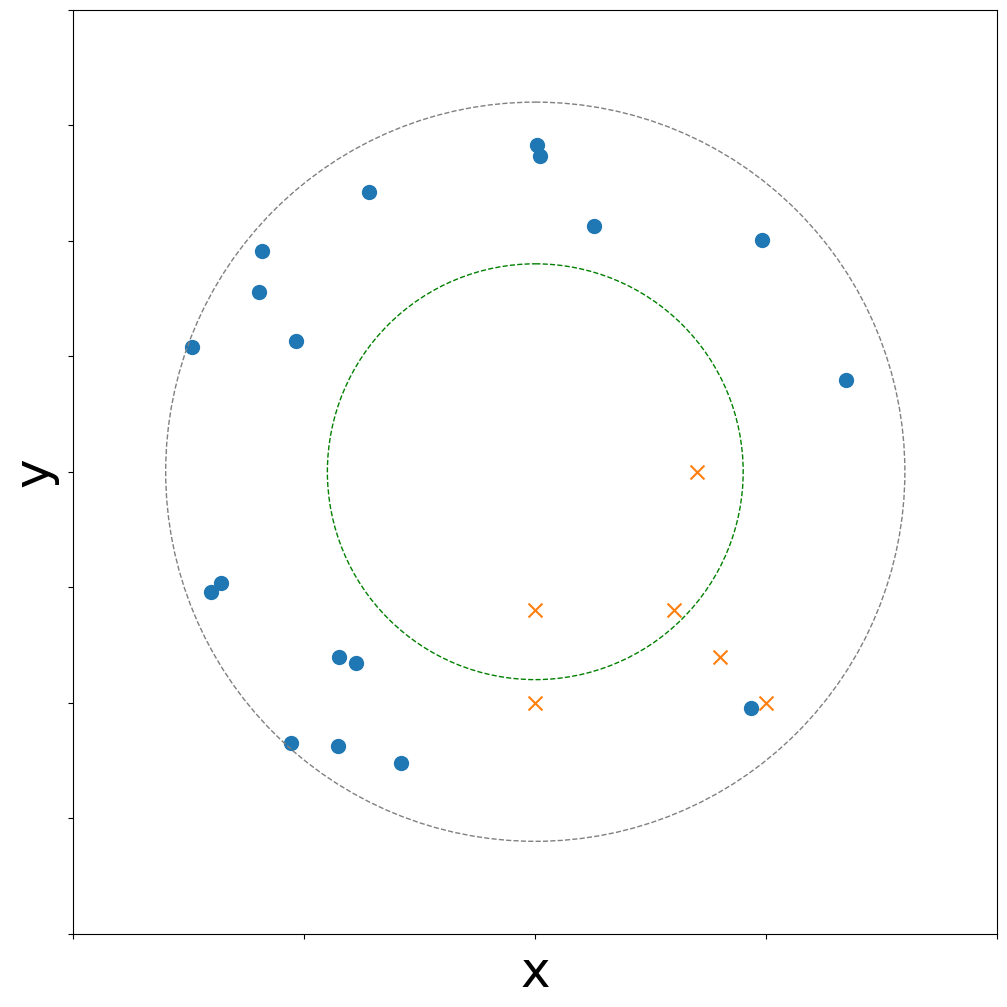}
        \caption{Goal locations from Section \ref{sec:experiments} and new test goals for evaluating PPO capability.}
        \label{fig:ablation-maze}
    \end{subfigure}
    \hfill
    \begin{subfigure}{0.4\linewidth}
        \centering
        \begin{tikzpicture}
    \node[draw=black, rounded corners, inner sep=2pt, fill=white] (legend) at (0,0) {
        \begin{tikzpicture}[scale=0.7]
            \def\linelen{0.3}
            \def\textgap{0.1}
            \def\yA{0}
            \def\yB{-0.5}

            \draw[thick, color={rgb,255:red,85; green,168; blue,104}, opacity=0.8] (0,\yA) -- ({\linelen},\yA);
            \fill[color={rgb,255:red,85; green,168; blue,104}, opacity=0.2] (0,{\yA - 0.1}) rectangle ({\linelen},{\yA + 0.1});
            \node[anchor=west, font=\scriptsize] at ({\linelen + \textgap}, \yA) {$|goal| < r$ (inside)};

            \draw[thick, color=gray, opacity=0.8] (0,\yB) -- ({\linelen},\yB);
            \fill[color=gray, opacity=0.2] (0,{\yB - 0.1}) rectangle ({\linelen},{\yB + 0.1});
            \node[anchor=west, font=\scriptsize] at ({\linelen + \textgap}, \yB) {$|goal| > r$ (outside)};
        \end{tikzpicture}
    };
\end{tikzpicture}
        \centering
        \includegraphics[width=\linewidth]{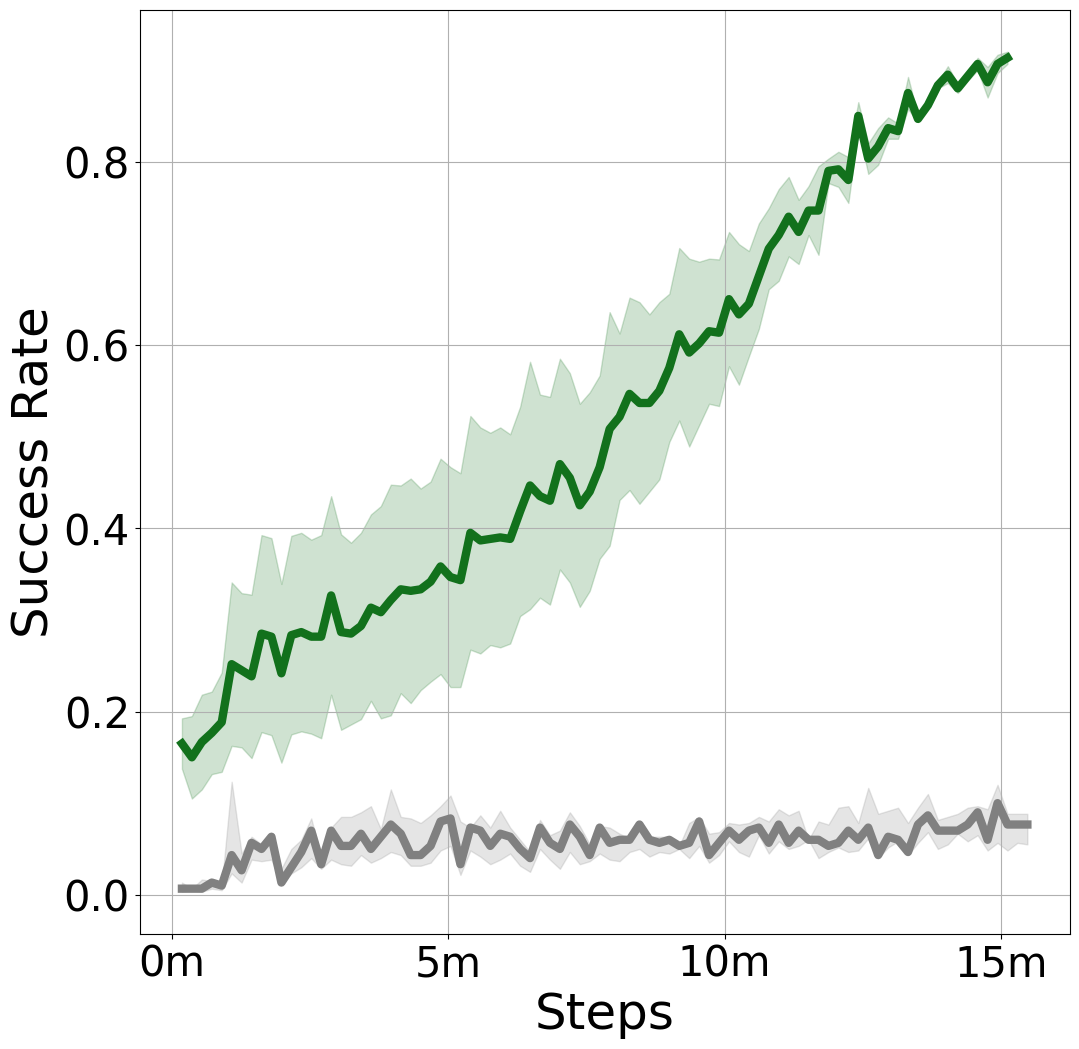}
        \caption{Median ± IQR success rate across $6$ runs for PPO without pretraining} 
        \label{fig:ablation-maze}
    \end{subfigure}
    \caption{    
        Comparison between goal region visualization and success rate performance for non-pretrained PPO agents.
        The left plot shows the spatial goal distribution with circular thresholds of radius $r$ where PPO without training reaches a good perfomance,
        while the right plot reports the corresponding success rates over time.
    }
    \label{fig:goal-success-comparison}
\end{figure}
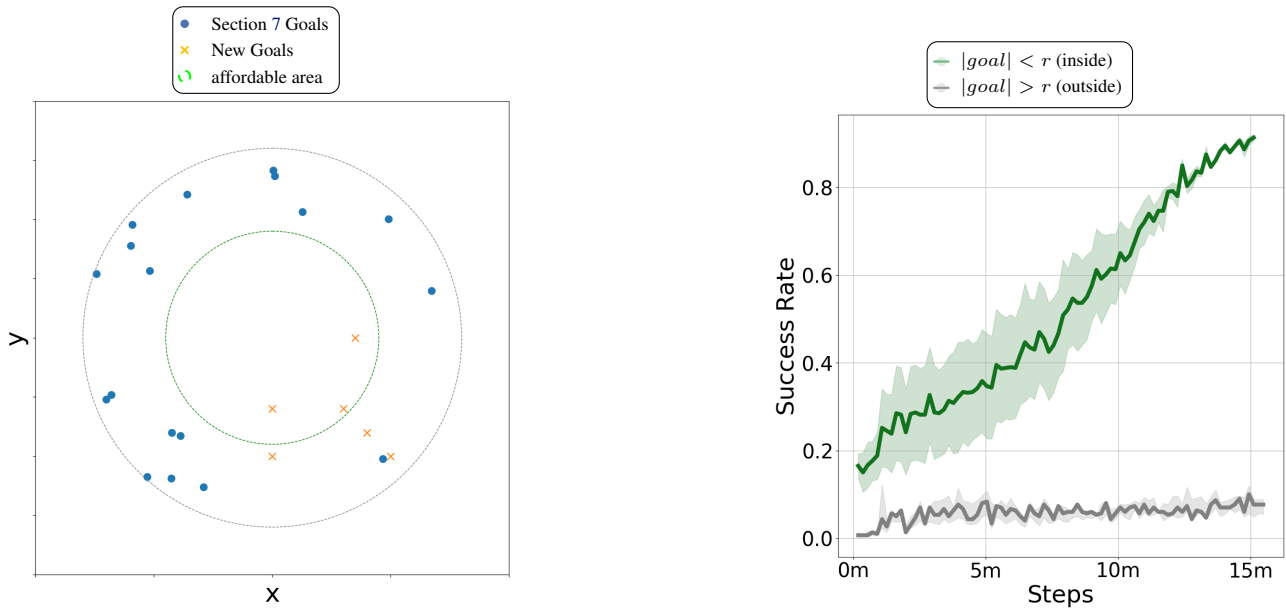

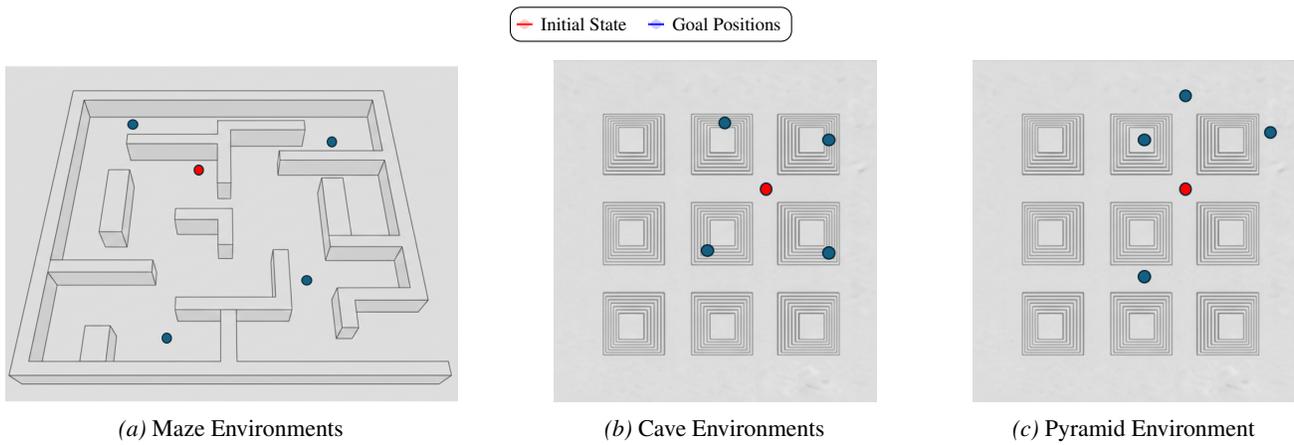
\begin{figure*}[ht]
    \centering

    \vspace{4pt}
    \begin{tikzpicture}
    \node[draw=black, rounded corners, inner sep=2pt, fill=white] (legend) at (0,0) {
        \begin{tikzpicture}[scale=0.7]
            \def\linelen{0.3}
            \def\textgap{0.05}
            \def\y{0}

            \def\xA{0}
            \def\xB{2.5}

            \draw[thick, color=red, opacity=0.8] (\xA,\y) -- ({\xA + \linelen},\y);
            \fill[color=red, opacity=0.2] (\xA,{\y - 0.1}) rectangle ({\xA + \linelen},{\y + 0.1});
            \node[anchor=west, font=\scriptsize] at ({\xA + \linelen + \textgap}, \y) {Initial State};

            \draw[thick, color=blue, opacity=0.8] (\xB,\y) -- ({\xB + \linelen},\y);
            \fill[color=blue, opacity=0.2] (\xB,{\y - 0.1}) rectangle ({\xB + \linelen},{\y + 0.1});
            \node[anchor=west, font=\scriptsize] at ({\xB + \linelen + \textgap}, \y) {Goal Positions};

        \end{tikzpicture}
    };
\end{tikzpicture}
    \vspace{6pt}

    \begin{subfigure}{0.35\linewidth}
        \centering
        \includegraphics[width=\linewidth]{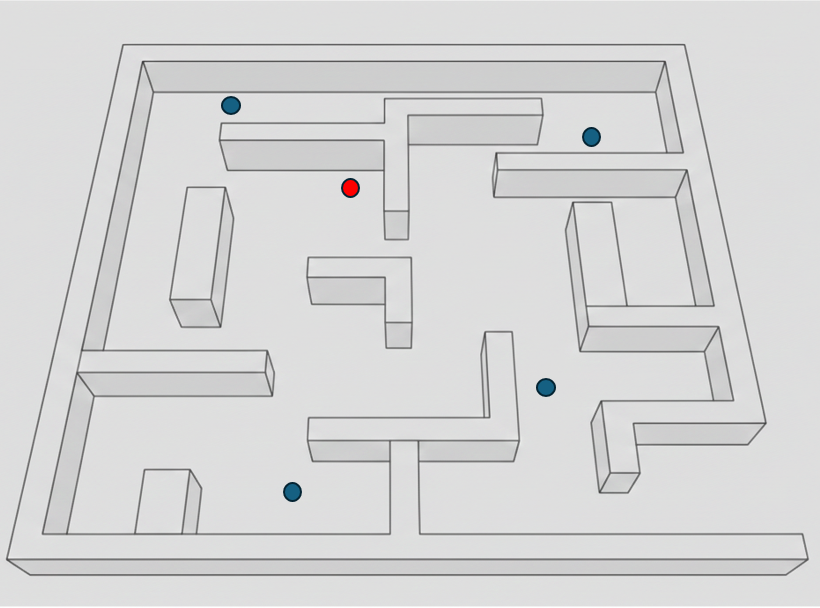}
        \caption{Maze Environments}
        \label{fig:maze-goals}
    \end{subfigure}
    \hfill
    \begin{subfigure}{0.25\linewidth}
        \centering
        \includegraphics[width=\linewidth]{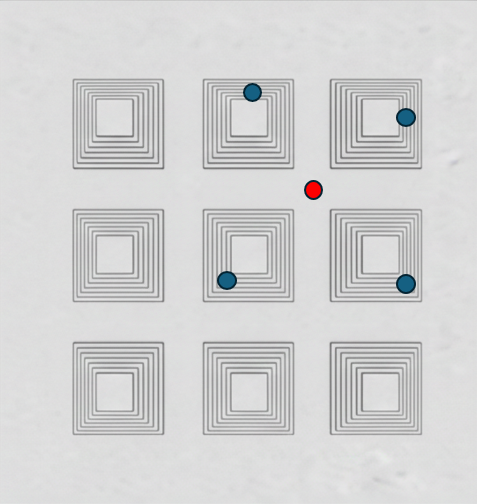}
        \caption{Cave Environments}
        \label{fig:cave-goals}
    \end{subfigure}
    \hfill
    \begin{subfigure}{0.25\linewidth}
        \centering
        \includegraphics[width=\linewidth]{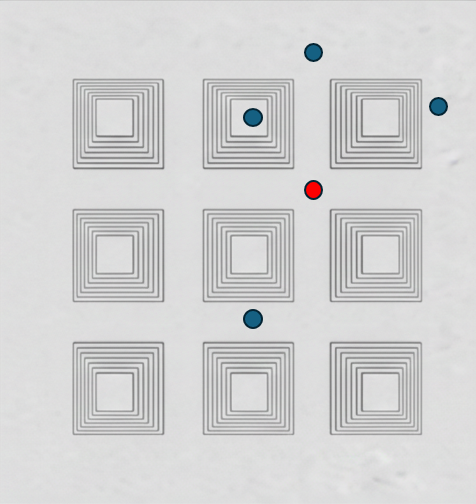}
        \caption{Pyramid Environment}
        \label{fig:pyramid-goals}
    \end{subfigure}

    \caption{
        Comparison between spatial goal distributions and PPO performance without pretraining.
        The left plot shows sampled goal regions with circular success thresholds of radius $r$,
        while the center and right plots report median success rates over time.
    }
    \label{fig:goal-success-comparison}
\end{figure*}

\begin{figure}[t]
  \centering
  \setlength{\tabcolsep}{2pt} 
  \renewcommand{\arraystretch}{0}%
  \begin{tabular}{ccc}
    \begin{subfigure}{0.30\textwidth}\centering
      \includegraphics[width=\linewidth]{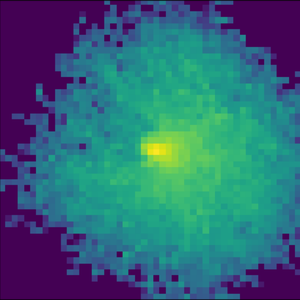}
    \end{subfigure} &
    \begin{subfigure}{0.30\textwidth}\centering
      \includegraphics[width=\linewidth]{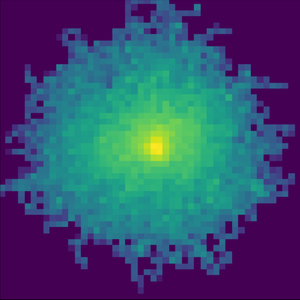}
    \end{subfigure} &
    \begin{subfigure}{0.30\textwidth}\centering
      \includegraphics[width=\linewidth]{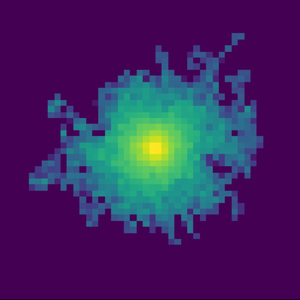}
    \end{subfigure} \\
    [4pt]\multicolumn{3}{c}{\small \textbf{Empty env}} \\[4pt]

    \begin{subfigure}{0.30\textwidth}\centering
      \includegraphics[width=\linewidth]{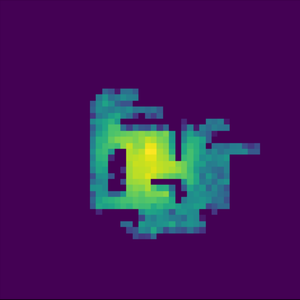}
    \end{subfigure} &
    \begin{subfigure}{0.30\textwidth}\centering
      \includegraphics[width=\linewidth]{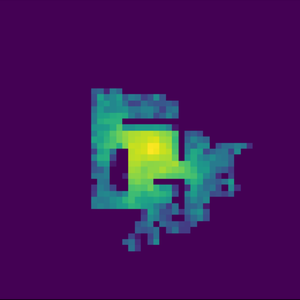}
    \end{subfigure} &
    \begin{subfigure}{0.30\textwidth}\centering
      \includegraphics[width=\linewidth]{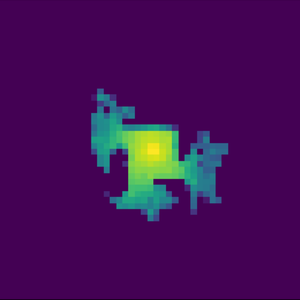}
    \end{subfigure} \\
    [4pt]\multicolumn{3}{c}{\small \textbf{Maze env}} \\[4pt]

    \begin{subfigure}{0.30\textwidth}\centering
      \includegraphics[width=\linewidth]{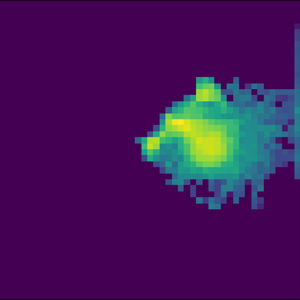}
    \end{subfigure} &
    \begin{subfigure}{0.30\textwidth}\centering
      \includegraphics[width=\linewidth]{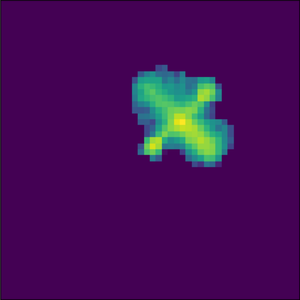}
    \end{subfigure} &
    \begin{subfigure}{0.30\textwidth}\centering
      \includegraphics[width=\linewidth]{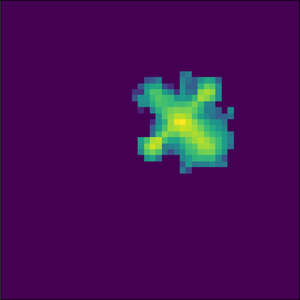}
    \end{subfigure} \\
    [4pt]\multicolumn{3}{c}{\small \textbf{Cave env}} \\

    \begin{subfigure}{0.30\textwidth}\centering
      \includegraphics[width=\linewidth]{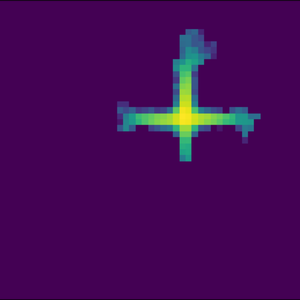}
    \end{subfigure} &
    \begin{subfigure}{0.30\textwidth}\centering
      \includegraphics[width=\linewidth]{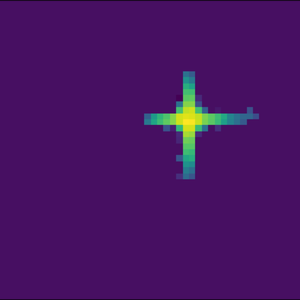}
    \end{subfigure} &
    \begin{subfigure}{0.30\textwidth}\centering
      \includegraphics[width=\linewidth]{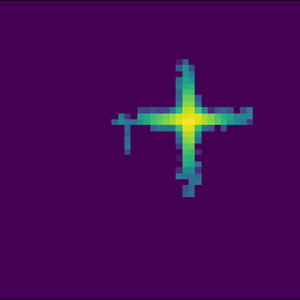}
    \end{subfigure} \\
    [4pt]\multicolumn{3}{c}{\small \textbf{Pyramid env}} \\[4pt]
  \end{tabular}

  \caption{2D heatmaps of the four environment configurations. Each row corresponds to a different environment, with columns showing the baseline, 10-agent, and 50-agent settings for the same random seed.}
  \label{fig:heatmaps}
\end{figure}

\end{document}